\documentclass[pdflatex,sn-mathphys-num]{sn-jnl}

\usepackage{float}	
\usepackage{graphicx}
\usepackage{url}
\usepackage{fullpage}
\usepackage{microtype}
\usepackage[utf8]{inputenc}
\usepackage[parfill]{parskip}
\usepackage{booktabs}
\usepackage{listings}
\usepackage[dvipsnames]{xcolor}
\usepackage[justification=centering]{caption}
\usepackage{textcomp,gensymb}
\usepackage{amsmath}
\usepackage{pdfpages}
\usepackage{subfiles}
\usepackage{blindtext}
\usepackage[colorinlistoftodos]{todonotes}
\usepackage{caption}
\usepackage{subcaption}
\usepackage{color,soul}
\usepackage{multirow}
\usepackage{multicol}
\usepackage{hhline}
\usepackage{comment}
\usepackage{titlesec}
\usepackage{amsmath}
\usepackage{geometry}
\usepackage{fancyhdr}      
\usepackage{marvosym}
\usepackage{wrapfig}
\usepackage{longtable}
\usepackage[super]{nth}

\usepackage{hyperref}
\usepackage[export]{adjustbox}

\usepackage{enumitem}
\usepackage{algorithm}
\usepackage{algorithmic}

\usepackage{array}   
\usepackage{longtable}  
\usepackage{parskip}  

\usepackage[english]{babel}

\usepackage{inconsolata}  

\definecolor{lightblue}{RGB}{240,245,255}  
\definecolor{mediumblue}{RGB}{51,119,255}  
\definecolor{darkblue}{RGB}{25,75,176}     
\definecolor{white}{RGB}{255,255,255}
\definecolor{correctgreen}{RGB}{240,248,244}
\definecolor{correctborder}{RGB}{70,136,71}
\definecolor{incorrectred}{RGB}{255,235,235}
\definecolor{incorrectborder}{RGB}{139,0,0}
\definecolor{separatorcolor}{RGB}{200,200,200}

\theoremstyle{thmstyleone}%
\theoremstyle{thmstyletwo}%

\theoremstyle{thmstylethree}%

\begin{document}

\title[KG-RAG for Question Answering]{Knowledge Graph-extended Retrieval Augmented Generation for Question Answering}


\author[1]{\fnm{Jasper} \sur{Linders}}\email{jasper.linders@gmail.com}

\author*[1]{\fnm{Jakub M.} \sur{Tomczak}}\email{jmk.tomczak@gmail.com\footnote{Currently working at Chan Zuckerberg Initiative: \url{jtomczak@chanzuckerberg.com}}}

\affil[1]{\orgdiv{Department of Mathematics and Computer Science}, \orgname{Eindhoven University of Technology}, \orgaddress{\street{De Zaale}, \city{Eindhoven}, \postcode{5600 MB}, \country{the Netherlands}}}


\abstract{Large Language Models (LLMs) and Knowledge Graphs (KGs) offer a promising approach to robust and explainable Question Answering (QA). While LLMs excel at natural language understanding, they suffer from knowledge gaps and hallucinations. KGs provide structured knowledge but lack natural language interaction. Ideally, an AI system should be both robust to missing facts as well as easy to communicate with. This paper proposes such a system that integrates LLMs and KGs without requiring training, ensuring adaptability across different KGs with minimal human effort. The resulting approach can be classified as a specific form of a Retrieval Augmented Generation (RAG) with a KG, thus, it is dubbed Knowledge Graph-extended Retrieval Augmented Generation (KG-RAG). It includes a question decomposition module to enhance multi-hop information retrieval and answer explainability. Using In-Context Learning (ICL) and Chain-of-Thought (CoT) prompting, it generates explicit reasoning chains processed separately to improve truthfulness. Experiments on the MetaQA benchmark show increased accuracy for multi-hop questions, though with a slight trade-off in single-hop performance compared to LLM with KG baselines. These findings demonstrate KG-RAG’s potential to improve transparency in QA by bridging unstructured language understanding with structured knowledge retrieval.}




\maketitle
\section{Introduction}
\label{sec:intro}

As our world becomes increasingly digital and information is more widely available than ever before, technologies that enable information retrieval and processing have become indispensable in both our personal and professional lives. The advent of Large Language Models (LLMs) has had a great impact, by changing the way many internet users interact with information, through models like ChatGPT\footnote{\url{https://chatgpt.com/}}. This has arguably played a large role in sparking an immense interest in solutions that build on artificial intelligence.

The rapid adoption of LLMs has transformed the fields of natural language processing (NLP) and information retrieval (IR). Understanding of natural language, with its long range dependencies and contextual meanings, as well as human-like text generation capabilities, allows these models to be applied to a wide variety of tasks. Additionally, LLMs have proven to be few-shot learners, meaning that they have the ability to perform unseen tasks with only a couple of examples \cite{Brown2020}. Unfortunately, the benefits of LLMs come at the cost of characteristic downsides, which are important to consider.

LLMs can hallucinate \cite{Ji2023a}, generating untruthful or incoherent outputs. They also miss knowledge not present during training, leading to knowledge cutoff, and cannot guarantee that certain training data is remembered \cite{PanKalo2023}. Because of their massive size and data requirements, LLMs are expensive to train, deploy, and maintain \cite{Bender2021}. Thus, smaller models or those needing only fine-tuning can be more practical for many use cases.

By contrast, Knowledge Graphs (KGs) store information explicitly as entities and relationships, allowing symbolic reasoning and accurate answers \cite{Pan2023}. Even if a direct link between entities is missing, inferences can be drawn from their shared associations. KGs may also recall underrepresented knowledge better than LLMs \cite{PanKalo2023}. However, they are costly to build, specialized to a domain, and typically require querying languages rather than natural language \cite{Yang2023}. They also do not easily generalize to other domains \cite{Pan2023}.

\begin{figure}[h!]
    \centering
    \includegraphics[width=1\linewidth]{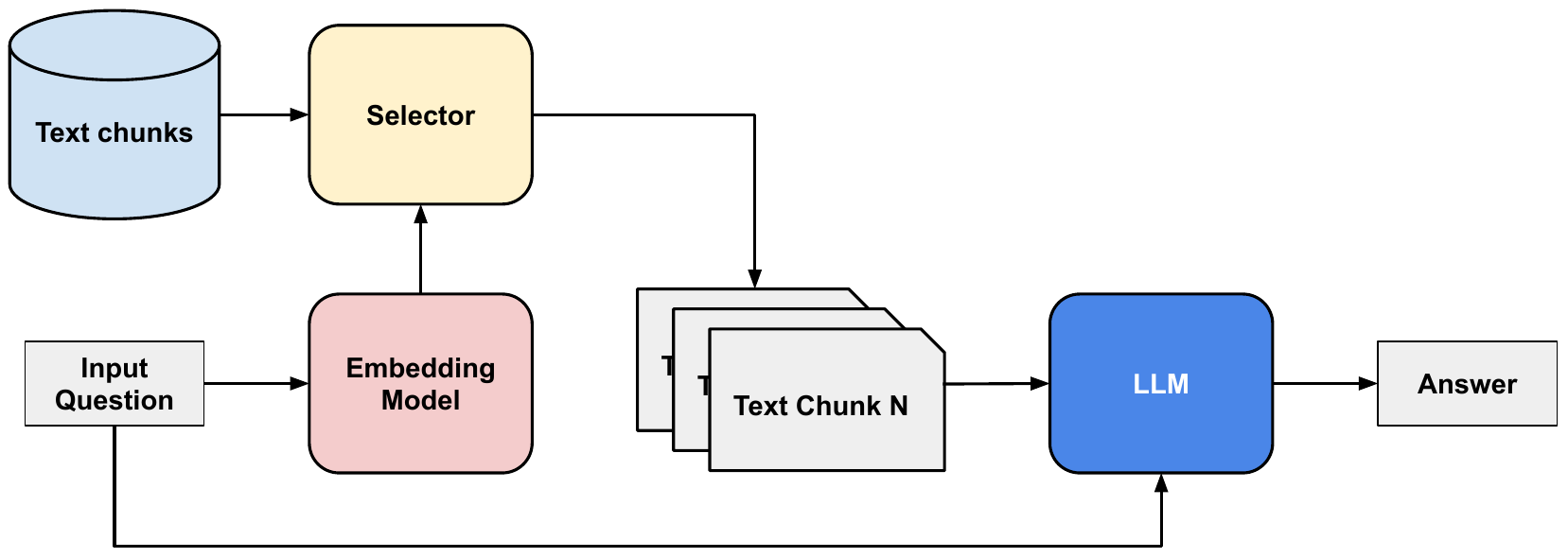}
    \caption{An example of a Retrieval-Augmented Generation (RAG) system, which combines information retrieval and text generation techniques. The red block indicates processing by a text embedding model, whereas the blue block depicts processing by an LLM. The yellow block shows a selector of nearest text chunks in the database.}
    \label{fig:RAG_example}
\end{figure}

Retrieval-Augmented Generation (RAG) \cite{Lewis2020} addresses LLMs’ lack of external knowledge by augmenting them with a text document database. Text documents are split into chunks, embedded, and stored in a vector database; the most similar chunks to an input query are retrieved and added to a prompt so the LLM can generate an answer based on this external information \cite{Gao2024} (see \autoref{fig:RAG_example}). However, relying on unstructured text can miss comprehensive entity data and even introduce distracting misinformation \cite{Gao2024}.

To overcome these limitations, RAGs can utilize KGs. The resulting system integrates structured data from Knowledge Graphs in a RAG, enabling precise retrieval and complex reasoning. For example, KAPING \cite{Baek2023} performs Knowledge Graph Question Answering (KGQA) without requiring any training. When training is needed for KG-enhanced LLMs, issues arise such as limited training data, domain specificity, and the need for frequent retraining as KGs evolve \cite{Wang2023, Wu2023}. In short, while RAG enhances LLMs by providing explainable, natural language outputs, incorporating structured Knowledge Graphs may offer improved reasoning and domain adaptability.

In this paper, we propose the Knowledge Graph-extended Retrieval Augmented Generation (KG-RAG) system, which combines the reliability of Retrieval Augmented Generation (RAG) with the high precision of Knowledge Graphs (KGs) and operates without any training or fine-tuning. We focus on the task of Knowledge Graph Question Answering; although this focus is narrow, our findings may have broader implications. For instance, certain insights could be applied to the development of other systems that utilize KG-based information retrieval, such as chatbots. The primary objective of this work is to investigate how LLMs can be enhanced through the integration of KGs. Since the term "enhance" can encompass various improvements, we define it as follows. First, we aim to enable LLMs to be more readily applied across different domains requiring specialized or proprietary knowledge. Second, we seek to improve answer explainability, thereby assisting end users in validating LLM outputs. Eventually, we aim to answer the following research questions:
\begin{enumerate} 
    \item How can Large Language Models be enhanced with Knowledge Graphs without requiring any training?
    \item How can answer explainability be improved with the use of Knowledge Graph-extended Retrieval Augmented Generation systems?    
\end{enumerate}

\section{Related Work}
\label{sec:related_work}

\paragraph{Knowledge Graphs} Knowledge Graphs (KGs) are structured databases that model real-world entities and their relationships as graphs, which makes them highly amenable to machine processing. They enable efficient querying to retrieve all entities related to a given entity, a task that would be significantly more challenging with unstructured text databases. Complex queries are executed using specialized languages such as SPARQL \cite{arenas2013}. As noted in recent research, "the success of KGs can largely be attributed to their ability to provide factual information about entities with high accuracy" \cite{PanKalo2023}. Typically, the information in KGs is stored as triples, i.e. $(subject, relation, object)$.

\paragraph{Large Language Models} Large Language Models (LLMs) learn natural language patterns from extensive text data, enabling various NLP tasks such as text generation and sentiment classification. Their emergence was enabled by the Transformer architecture, introduced in \textit{Attention Is All You Need} \cite{Vaswani2017}, which efficiently models sequential data via attention mechanisms. Scaling these models—by increasing compute, dataset size, and parameter count—yields performance improvements following a power law \cite{kaplan2020}, with LLMs typically comprising hundreds of millions to hundreds of billions of parameters.

LLMs generate text in an autoregressive manner. Given a sequence $x_{1:t}$, the model produces a probability distribution $p(x_{t+1}|x_{1:t}) = \mathrm{softmax}(z/T)$ over its vocabulary, where $z$ are the raw logits and $T$ is a temperature parameter that controls randomness. Instead of selecting tokens via simple $\mathrm{argmax}$, more sophisticated sampling methods are employed (see \autoref{sec:components}) to generate coherent and diverse output consistent with the input context \cite{Zhao2023}.

\paragraph{In-Context Learning \& Chain-of-Thought}

In-Context Learning (ICL) improves LLM performance by providing few-shot examples instead of zero-shot queries. This method boosts task performance through prompt engineering without altering model parameters \cite{chen2023}. It is often combined with Chain-of-Thought (CoT) that can significantly enhance performance without modifying the model's parameters or incurring the high cost of fine-tuning \cite{Wei2023}. A CoT prompt instructs the model to generate intermediate reasoning steps that culminate in the final answer, rather than directly mapping a query to an answer \cite{Wei2023}. This approach naturally decomposes complex queries into simpler steps, yielding more interpretable results.

\paragraph{Knowledge Graph Question Answering} Knowledge Graph Question Answering (KGQA) is the task of answering questions using a specific knowledge graph (KG). Benchmarks such as Mintaka \cite{Sen2022}, WebQuestionsSP \cite{Yih2016}, and MetaQA \cite{Zhang} provide datasets where each row includes a question, its associated entity/entities, and the answer entity/entities, along with the corresponding KG (provided as a file of triples or accessible via an API). In these benchmarks, the question entity is pre-identified (avoiding the need for entity matching or linking), and performance is evaluated using the binary Hit@1 metric.

KGQA systems are typically classified into three categories \cite{Baek2023}:
\begin{itemize}
    \item Neural Semantic Parsing-Based Methods: These map a question to a KG query (e.g., in SPARQL), reducing the search space between question and answer entities. Although effective \cite{Yih2016}, they require labor-intensive semantic parse labels.
    \item Differentiable KG-Based Methods: These employ differentiable representations of the KG (using sparse matrices for subjects, objects, and relations) to perform query execution in the embedding space. They enable end-to-end training on question-answer pairs \cite{Oliya2021, Sen2023}, but necessitate ample training data and may not generalize across different KGs.
    \item Information Retrieval-Based Methods: These combine KGs with LLMs by retrieving relevant facts—which are then injected into the prompt—to generate answers \cite{Baek2023}. Although they leverage off-the-shelf components, they often require fine-tuning on KG-specific datasets \cite{Wu2023}.
\end{itemize}

\paragraph{Knowledge Graph-extended Retrieval Augmented Generation} Information retrieval-based KGQA (IR-KGQA) systems differ from neural semantic parsing and differentiable KG methods by delegating part of the reasoning over triples to the LLM. The process is split into retrieving candidate triples and then having the LLM reason over them to formulate an answer, whereas the other methods map directly from the question to the answer entities \cite{Oliya2021, Gu2022}.

KG-RAG is defined as an IR-KGQA system that employs a similarity-based retrieval mechanism using off-the-shelf text embedding models, akin to the original RAG system \cite{Lewis2020}. In KG-RAG (exemplified by the KAPING system \cite{Baek2023}), candidate triples are retrieved up to $N$ hops from the question entity/entities, verbalized, and embedded alongside the question. Their similarity is computed via dot or cosine product, and the Top-$K$ similar triples are passed to an answer generation LLM, which then outputs the answer.

\section{Methodology}
\label{sec:methodology}

\subsection{Problem Statement}
\label{sec:formal_problem}
Let $G$ be a knowledge graph, defined as a set of triples of the form $(s, r, o)$ where:
\begin{itemize}
    \item Each triple $(s, r, o) \in G \subseteq \mathcal{E} \times \mathcal{R} \times \mathcal{E}$ represents a fact;
    \item $s, o \in \mathcal{E}$ are entities from the set of all entities $\mathcal{E}$;
    \item $r \in \mathcal{R}$ is a relation from the set of all relations $\mathcal{R}$.
\end{itemize}
We assume that the following objects are given:
\begin{itemize}
    \item A question $q$ that can be answered using facts from $G$
    \item The question entity/entities part of that question $e_q \in \mathcal{E}$
\end{itemize}
Moreover, let us introduce the following variables:
\begin{itemize}
    \item $a$ denotes a natural language answer that can be derived from the facts in $G$;
    \item $c$ is a reasoning chain in natural language, explaining the \textit{logical} steps from $q$ and $e_q$ to $a$
\end{itemize}

\vspace{4mm}

Our \textbf{objective} is to develop a function $f$ that maps given object to both an answer and the reasoning chain, namely:

\[f: q \times e_q \times G \rightarrow (a, c)\]

where:
\begin{itemize}
    \item $a$ is a natural language answer that can be derived from the facts in $G$
    \item $c$ is a reasoning chain in natural language, explaining the logical steps from $q$ and $e_q$ to $a$
\end{itemize}

\vspace{4mm}

Additionally, we aim for the following:
\begin{itemize}
    \item \textbf{Answer Accuracy:} The function $f$ should have high answer accuracy, as evaluated by the Hit@1 metric.
    \item \textbf{Answer Explainability:} For each answer $a$ generated by the function $f$, the reasoning chain $c$ must provide a clear logical explanation of how the answer was derived, so that it is more easily verifiable by the user.
    \item \textbf{Application Generalizability:} The function $f$ must operate without training or finetuning on specific Knowledge Graphs, using only In-Context Learning examples. The Knowledge Graphs must include sufficient amounts of natural language information, as the system relies on natural language-based methods.
\end{itemize}

The degree to which the function $f$ achieves the objectives is evaluated using both quantitative and qualitative methods, based on experiments with a KGQA benchmark, namely:
\begin{itemize}
    \item Quantitative evaluation of answer accuracy, based on the Hit@1 metric.
    \item Qualitative analysis of reasoning chain clarity and logical soundness, as judged by a human evaluator on a sample of results.
\end{itemize}

\subsection{State-of-the-Art}\label{sec:models}

Recent advances in question answering have seen the development of several state-of-the-art methods that leverage a diverse array of Large Language Models alongside innovative baseline strategies. For instance, one method employs multiple scales of models such as T5, T0, OPT, and GPT-3, while experimenting with baselines ranging from no knowledge to generated knowledge on datasets like WebQSP \cite{Yih2016} and Mintaka \cite{Sen2022}. Another approach expands this exploration by integrating Llama-2, Flan-T5, and ChatGPT, and introducing baselines that utilize triple-form knowledge and alternative KG-to-Text techniques, evaluated on datasets that include WebQSP, MetaQA \cite{Zhang}, and even a Chinese benchmark, ZJQA \cite{Wu2023}. Additionally, methods centered on ChatGPT are further compared with systems like StructGPT and KB-BINDER across varying complexities of MetaQA and WebQSP. The overview of the SOTA methods is presented in \autoref{tab:models}.

\begin{table}[h!]
\small
\centering
\caption{Comparison of the question answering LLMs, baselines and benchmark datasets that were used for the different models. The full set of QA LMs is as follows: T0 \cite{Sanh2022}, T5 \cite{Raffel2020}, Flan-T5 \cite{Chung2022}, OPT \cite{Zhang2022}, GPT-3 \cite{Brown2020}, ChatGPT, AlexaTM \cite{Fitzgerald2022}, and Llama-2 \cite{Touvron2023}. The full set of datasets is as follows: WebQuestions \cite{Berant2013}, WebQSP \cite{Yih2016}, ComplexWebQuestions \cite{Talmor2013}, MetaQA \cite{Zhang}, Mintaka \cite{Sen2022}, LC-QuAD \cite{Dubey}, and ZJQA \cite{Wu2023}.\\}
\label{tab:models}
\begin{tabular}{l|lll}
\textbf{Model} & \textbf{QA LMs} & \textbf{Baselines} & \textbf{Datasets} \\ \hline

KAPING \cite{Baek2023} 
& \begin{tabular}[c]{@{}l@{}}T5 (0.8B, 3B, 11B)\\ T0 (3B, 11B)\\ OPT (2.7B, 6.7B)\\ GPT-3 (6.7B, 175B)  \end{tabular}            
& \begin{tabular}[c]{@{}l@{}}No knowledge\\ Random knowledge\\ Popular knowledge\\ Generated knowledge\end{tabular}                                      & \begin{tabular}[c]{@{}l@{}}WebQSP (w/ 2 KGs)\\ Mintaka \end{tabular}\\ \hline

Retrieve-Rewrite-Answer \cite{Wu2023}     
& \begin{tabular}[c]{@{}l@{}}Llama-2 (7B, 13B)\\ T5 (0.8B, 3B, 11B)\\ Flan-T5 (80M, 3B, 11B)\\ T0 (3B, 11B)\\ ChatGPT\end{tabular} 
& \begin{tabular}[c]{@{}l@{}}No knowledge\\ Triple-form knowledge\\ 2x Alternative KG-to-Text\\ 2x Rival model\end{tabular} 
& \begin{tabular}[c]{@{}l@{}}WebQSP\\ WebQ\\ MetaQA\\ ZJQA (Chinese)\end{tabular}        \\ \hline

Keqing \cite{Wang2023}   
& ChatGPT 
& \begin{tabular}[c]{@{}l@{}}ChatGPT\\ StructGPT\\ KB-BINDER\end{tabular} 
& \begin{tabular}[c]{@{}l@{}}WebQSP\\ MetaQA-1hop\\ MetaQA-2hop\\ MetaQA-3hop\end{tabular}  \end{tabular}
\end{table}

\subsubsection{KAPING}

KAPING \cite{Baek2023} is one of the best IR-KGQA models that requires no training. For example, due to the large number of candidate triples--27$\%$ of entities in WebQSP \cite{Yih2016} have more than 1000 triples--a text embedding-based selection mechanism is employed, typically using cosine similarity \cite{Pedersen2004}, instead of appending all triples directly to the prompt. KAPING outperforms many baselines presented in \autoref{tab:models} in terms of Hit@1, especially those with smaller LLMs, suggesting that external knowledge compensates for the limited parameter space. Notably, using 2-hop triples degrades performance, so only 1-hop triples are selected; when retrieval fails to fetch relevant triples, performance drops below a no-knowledge baseline. An additional finding is that triple-form text outperforms free-form text for retrieval, as converting triples to free-form via a KG-to-Text model often leads to semantic incoherence, and using free-form text in prompts does not improve answer generation.

\subsubsection{Retrieve-Rewrite-Answer}

Motivated by KAPING’s limitations, the Retrieve-Rewrite-Answer (RRA) architecture was developed for KGQA \cite{Wu2023}. Unlike KAPING, which overlooked the impact of triple formatting, RRA introduces a novel triple verbalization module, among other changes. Specifically, question entities are extracted from annotated datasets (with entity matching deferred). The retrieval process consists of three steps: (i) a hop number is predicted via a classification task on the question embedding; (ii) relation paths--sequences of KG relationships--are predicted by sampling and selecting the top-$K$ candidates based on total probability; (iii) selected relation paths are transformed into free-form text using a fine-tuned LLM. This verbalized output, together with the question, is fed to a QA LLM via a prompt template.

For training, the hop number and relation path classifiers, as well as the KG-to-Text LLM, are tuned on each benchmark. Due to the lack of relation path labels and subgraph-text pairs in most benchmarks, the authors employ various data construction techniques, limiting the model’s generalizability across domains and KGs.

As detailed in \autoref{tab:models}, evaluations were carried out using QA LLM, baselines (no knowledge, triple-form knowledge and two standard KG-to-Text models), and benchmark datasets, compared with models from \cite{Baek2023} and \cite{Sen2023} on WebQ \cite{Berant2013} and WebQSP \cite{Yih2016} using the Hit@1 metric. The main results show that RRA significantly outperforms rival models, achieving an improvement of 1–8\% over triple-form text and 1–5\% over the best standard KG-to-Text model. Moreover, RRA is about 100$\times$ more likely to produce a correct answer when the no-knowledge baseline fails, confirming the added value of IR-based KGQA models over vanilla LLMs.

\subsubsection{Keqing}
\label{sec:keqing}

Keqing, proposed in \cite{Wang2023}, is the third SOTA model that is positioned as an alternative to SQL-based retrieval systems. Its key innovation is a question decomposition module that uses a fine-tuned LLM to break a question into sub-questions. These subquestions are matched to predefined templates via cosine similarity, with each template linked to specific KG relation paths. Candidate triples are retrieved based on these relation paths, and sub-questions are answered sequentially--the answer to one sub-question seeds the next. The triples obtained are verbalized and processed through a prompt template by a Quality Assurance LLM, ultimately generating a final answer that reflects the model’s reasoning chain.

In this approach, only the question decomposition LLM is trained using LoRA \cite{Shen2021}, which adds only a small fraction of trainable weights. However, the construction of sub-question templates and the acquisition of relation path labels are not clearly detailed, which may limit the system's scalability.

According to \autoref{tab:models}, Keqing outperforms vanilla ChatGPT and two rival models, achieving Hit@1 scores of 98.4\% to 99.9\% on the MetaQA benchmark and superior performance on the WebQSP benchmark. Its ability to clearly explain its reasoning through sub-question chains further underscores its contribution to answer explainability.

\subsubsection{Research Gap}

After KAPING was introduced as the first KG-Augmented LLM for KGQA, RRA \cite{Wu2023} and Keqing \cite{Wang2023} followed, each employing different triple retrieval methods. Although all three use an LLM for question answering, KAPING relies on an untrained similarity-based retriever, while RRA and Keqing develop trainable retrieval modules, improving performance at the cost of significant engineering. Specifically, RRA trains separate modules (hop number classifier, relation path classifier, and KG-to-Text LLM) for each benchmark, requiring two custom training datasets (one for questions with relation path labels and one for triples with free-form text labels). The need for KG-specific techniques limits generalizability and raises concerns about the extra labor required when no Q\&A dataset is available. Keqing fine-tunes an LLM for question decomposition to enhance answer interpretability and triple retrieval. This approach also demands a training dataset with sub-question templates and relation path labels, though the methods for constructing these remain unclear. Consequently, it is debatable whether the performance gains justify the additional engineering effort.

In summary, these shortcomings reveal a gap for models that are both as generalizable as KAPING and as explainable as Keqing. KAPING’s training-free design allows minimal human intervention across diverse KGs and domains, even in the absence of benchmark datasets. For this reason, we propose an improvement to the KAPING model by introducing a question decomposition module.

\subsection{Our Approach}
\label{sec:proposed_solution}

KAPING, a SOTA method combinining KGs and LLMs, outperforms many zero-shot baselines. However, its retrieval process, a vital process for accurate answer generation, can benefit from reducing irrelevant triple inclusion \cite{Baek2023}. Therefore, we build on top of the KAPING model and propose to enhance it by integrating a question decomposition module to improve triple retrieval, answer accuracy, and explainability while maintaining application generalizability. 

The proposed question decomposition module decomposes complex, multi-hop questions into simpler sub-questions. This allows the similarity-based retriever to focus on smaller, manageable pieces of information, thereby improving retrieval precision and yielding a more interpretable reasoning chain. Unlike conventional Chain-of-Thought prompting, which may induce hallucinated reasoning \cite{Radha2023}, decomposing the question forces the LLM to independently resolve each sub-question, ensuring fidelity to the stated reasoning. Our question decomposition module uses manually curated in-context learning examples for the KGQA benchmark, obviating the need for additional training and minimizing human labor. As a result, our approach aligns well with the goals of enhanced generalizability and answer explainability while potentially outperforming KAPING for multi-hop questions. The following section details the overall system architecture and the roles of its individual components.

\subsection{System Architecture}\label{sec:architecture}

Our system comprises multiple components, each executing a specific role in answering KG-based questions. The overall process involves four primary steps, with the first two being non-sequential:

\begin{enumerate}
    \item \textbf{Question Decomposition:} The decomposition module splits the question into sub-questions. For simple queries, it avoids unnecessary decomposition.
    \item \textbf{Candidate Triple Retrieval:} Given the question entity, the system retrieves all triples up to $N$ hops from the KG. Each triple is verbalized into text for subsequent selection via a sentence embedding model.
    \item \textbf{Sub-Question Answering:} This sequential step answers each sub-question using the candidate triples. The process involves embedding the candidate triples to form a vector database, selecting the Top-$K$ similar triples for the sub-question, and reformulating subsequent sub-questions based on prior sub-answers.
    \item \textbf{Answer Synthesis:} Finally, the system synthesizes the final answer from the sub-questions and their corresponding answers. The output also includes the chain-of-thought from the decomposition stage, enhancing interpretability.
\end{enumerate}

\begin{figure}[]
    \centering
    \includegraphics[width=1\linewidth]{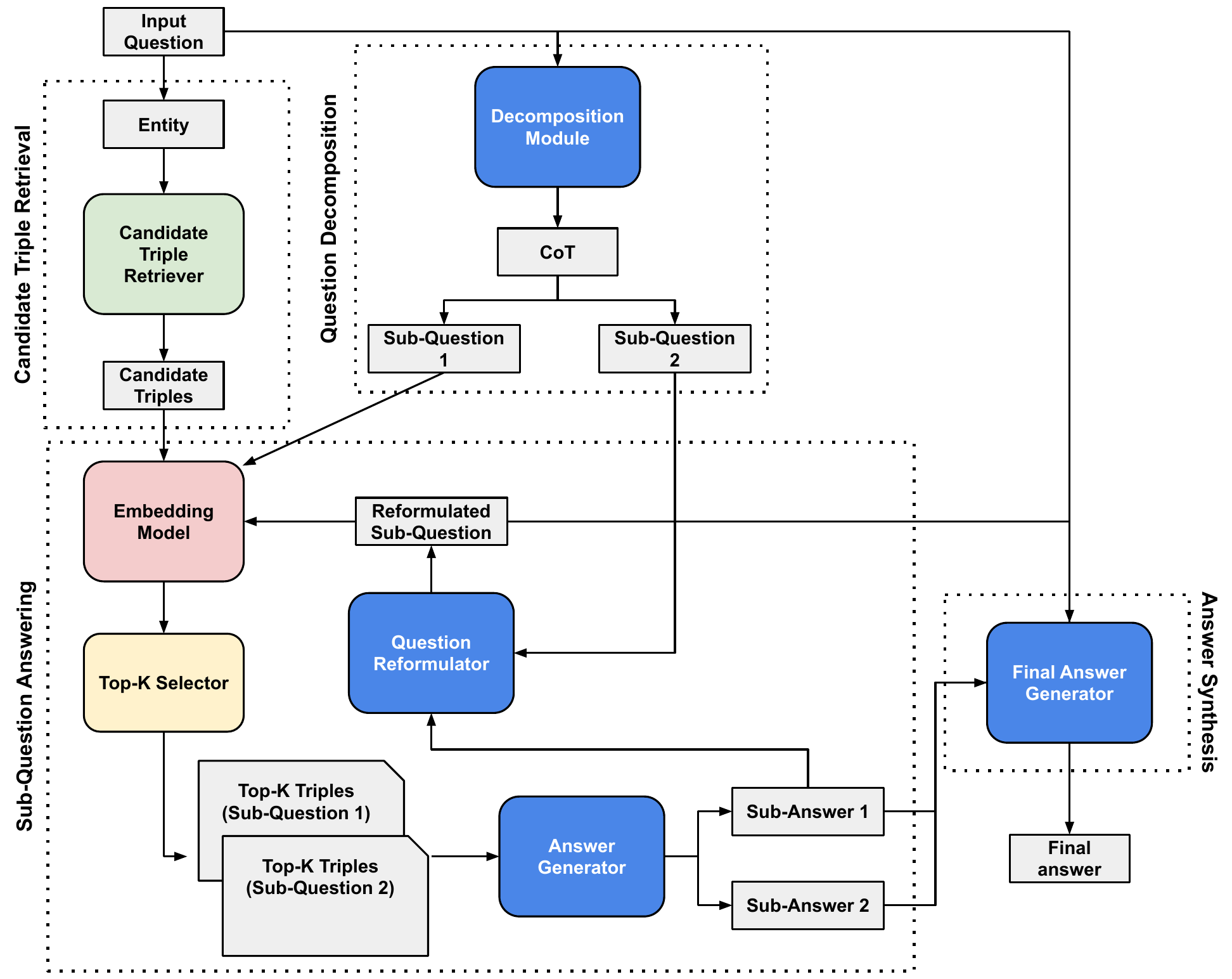}
    \caption{The architecture of the proposed system. An example of a 2-hop question is included, to give an idea of the data structures that are involved in the end-to-end process. The green color indicates processing with the KG; the red block shows the embedding model and the blue modules utilize an LLM.}
    \label{fig:system_architecture}
\end{figure}

\autoref{fig:system_architecture} illustrates the system architecture, highlighting the data structures and interactions between components. The diagram shows how the question reformulation module, which processes all previous sub-answers, enables the sequential resolution of sub-questions until the final answer is generated by the answer synthesis module.

Different components utilize distinct data sources and models. The candidate triple retriever directly accesses the KG, while the similarity-based triple selection leverages an off-the-shelf sentence embedding model trained on question-answer pairs. The remaining modules—the decomposition module, sub-answer generator, question reformulator, and final answer generator—are implemented using a LLM.

\subsection{System Components}\label{sec:components}

\subsubsection{Question Decomposition}

\paragraph{Overview}  
The question decomposition module splits a complex question into simpler sub-questions while generating an explicit reasoning chain, thereby enhancing both triple retrieval and answer explainability (\autoref{sec:proposed_solution}). Inspired by Chain-of-Thought and In-Context Learning techniques \cite{Radha2023}, the module uses manually constructed ICL examples from the benchmark (\autoref{sec:benchmark}). The prompt is designed to first elicit the reasoning chain (CoT) followed by the sub-questions, aligning with the natural text-based reasoning of LLMs.

\paragraph{Inputs and Outputs}  
As illustrated in \autoref{fig:system_architecture}, the module takes a natural language question as input and outputs a string containing the reasoning chain and sub-questions. This output is post-processed to extract the CoT and store the sub-questions in a list.

\paragraph{Techniques}  
The decomposition prompt instructs the LLM to decide if a question requires decomposition. If so, it generates a CoT followed by sub-questions, strictly adhering to a specified format and avoiding irrelevant content. In-context examples--covering three question types from the MetaQA benchmark--guide the LLM, with the stop token ``$<$END$>$'' marking completion.

\paragraph{Implementation Details}  
Here, we use a 4-bit quantized version of Mistral-7B-Instruct-v0.2 \cite{Touvron2023, Jiang2023}, originally a 7.24B-parameter model that outperforms Llama 2 and Llama 1 in reasoning, mathematics, and code generation. The quantized model, sized at 4.37 GB\footnote{\url{https://huggingface.co/TheBloke/Mistral-7B-Instruct-v0.2-GGUF}}, is compatible with consumer-grade hardware (e.g., NVIDIA RTX 3060 12GB\footnote{\url{https://www.msi.com/Graphics-Card/GeForce-RTX-3060-VENTUS-2X-12G-OC}}). Fast inference is achieved using the \texttt{llama.cpp} package\footnote{\url{https://github.com/ggerganov/llama.cpp}}, and prompts are designed with LM Studio\footnote{\url{https://lmstudio.ai/}}.

Inference parameters (see \autoref{tab:QD_LLM_parameters}) include a max tokens limit (256) to prevent runaway generation, a temperature of 0.3 to reduce randomness, and top-k (40) and min-p (0.05) settings to ensure controlled token sampling \cite{min_p}.

\begin{table}[h!]
\centering
\caption{The inference parameters that were used for the question decomposition LLM.}
\label{tab:QD_LLM_parameters}
\begin{tabular}{l|l}
\textbf{Parameter} & \textbf{Value} \\ \hline
Max Tokens         & 256            \\
Temperature        & 0.3            \\
Min-p              & 0.05           \\
Top-k              & 40            
\end{tabular}
\end{table}

\subsubsection{Candidate Triple Retrieval}
\paragraph{Overview}  
Candidate triple retrieval collects all triples up to $N$ hops from a given question entity in the KG, converting each triple into a text string of the form $(subject, relation, object)$. Although the worst-case complexity is exponential in the number of hops—approximately $\Theta(d^N)$ for an undirected KG with average degree $d$—real-world KGs are sparse, making the average or median complexity more relevant (\autoref{sec:benchmark}). The value of $N$ is treated as a hyperparameter.

\paragraph{Inputs and Outputs}  
This component accepts the question entity/entities as a natural language string and retrieves candidate triples from the KG. The output is a list of lists, where each sub-list corresponds to the candidate triples for each hop up to $N$. Each triple is stored as a formatted text string, with underscores replaced by spaces (e.g., "acted\_in" becomes "acted in").

\begin{figure}[H]
    \centering
    \includegraphics[width=0.8\linewidth]{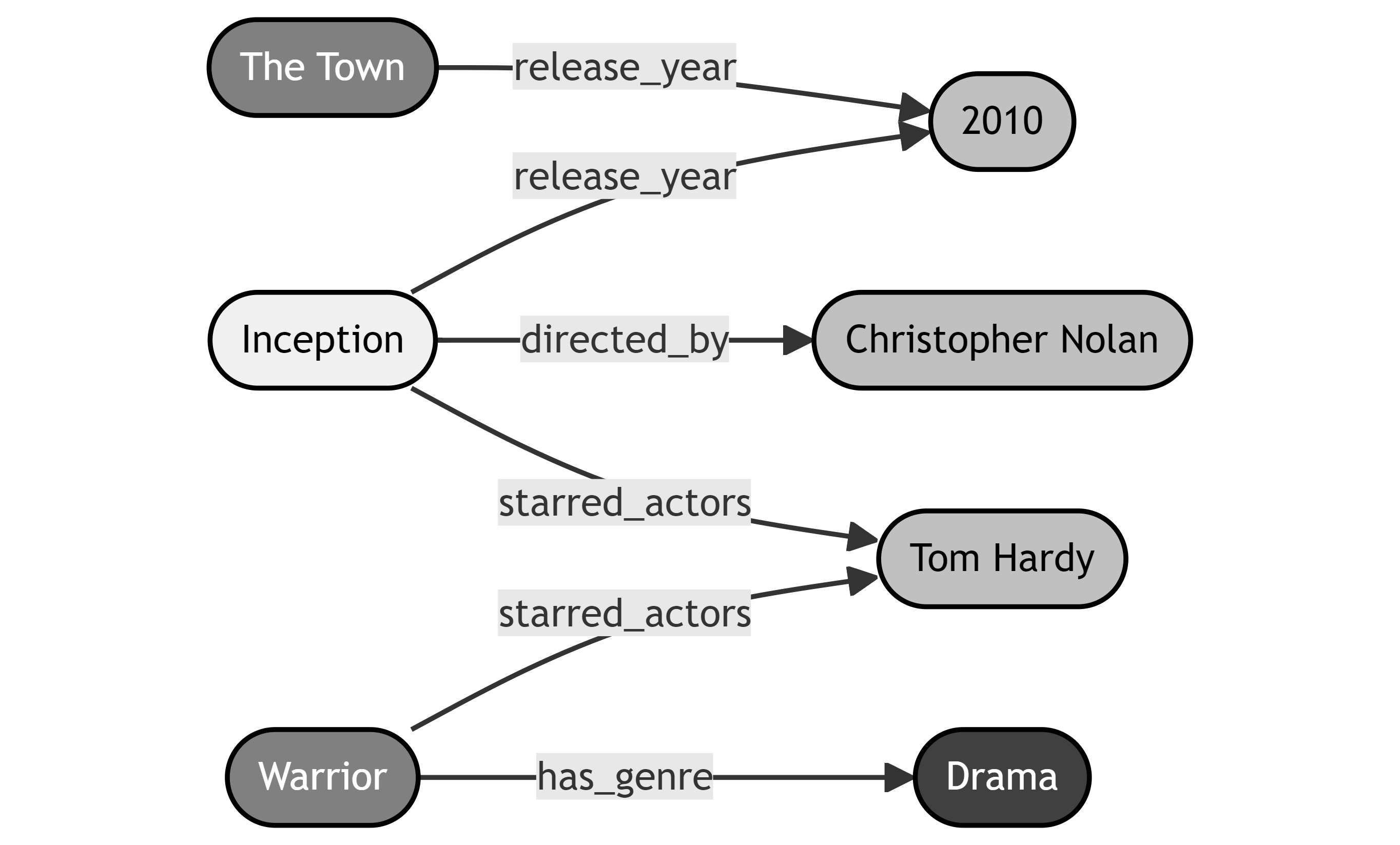}
    \caption{A simple subgraph of triples from MetaQA \cite{Zhang}. As indicated by the arrows, this KG is a directed graph, which has implications for candidate triple retrieval. If \textit{Inception} were the entity we were retrieving for, each darker tint of gray shows the entities that would be reached for a hop deeper.}
    \label{fig:KG_example}
\end{figure}

\paragraph{Techniques}  
Candidate triple retrieval employs a breadth-first search strategy. In the MetaQA benchmark, which uses a directed KG, retrieval can be unidirectional (considering only outgoing edges) or bidirectional (including both outgoing and incoming edges). For example, as illustrated in \autoref{fig:KG_example}, unidirectional retrieval from the \textit{Inception} entity would only yield entities like \textit{2010}, \textit{Christopher Nolan}, and \textit{Tom Hardy}, whereas bidirectional retrieval expands the search across successive hops. This example underscores the impact of retrieval direction on both the candidate set and computational load.

\paragraph{Implementation Details}  
The MetaQA benchmark provides the KG as a text file with one triple per row. This file is pre-processed into a compressed KG with indexed entities and relationships to streamline retrieval and minimize memory usage. Each triple is embedded using a sentence embedding model (introduced in \autoref{sec:sub_QA}), forming a dictionary of embeddings that enhances retrieval efficiency by avoiding redundant computations. Retrieval is performed bidirectionally up to 3 hops, i.e., $N \in \{1, 2, 3\}$.

\subsubsection{Sub-Question Answering}
\label{sec:sub_QA}

\paragraph{Overview}  
Once the question is decomposed into sub-questions and candidate triples are retrieved for the given entity/entities, the sub-question answering process begins. Iteratively, the sub-question and candidate triples are embedded using a sentence embedding model, and the top-$K$ similar triples are selected to generate a sub-answer via an LLM. This sub-answer is then used to reformulate subsequent sub-questions if needed (see \autoref{fig:system_architecture}), continuing until all sub-questions are answered.

\paragraph{Inputs and Outputs}  
Inputs include candidate triples (a list of strings, pre-embedded from the MetaQA KG) and a list of sub-questions. The output comprises two lists of strings: one containing the sub-answers and another with the reformulated sub-questions, both of which contribute to the final answer synthesis.

\paragraph{Techniques}  
The process employs similarity-based retrieval where both the sub-question and candidate triples are embedded with the same model, and their dot-product similarity is computed. The top-$K$ triples are then passed to a zero-shot LLM answer generator along with the sub-question. Unlike Keqing’s multiple-choice approach \cite{Wang2023} (\autoref{sec:keqing}), this method allows the LLM to reason over the context. A zero-shot LLM also performs question reformulation.

\paragraph{Implementation Details}  
The similarity-based triple selection uses the \texttt{multi-qa-mpnet-base-dot-v1}\footnote{\url{https://huggingface.co/sentence-transformers/multi-qa-mpnet-base-dot-v1}} model from the \texttt{sentence\_transformers}\footnote{\url{https://www.sbert.net/}} package, which embeds text into 768-dimensional vectors. Similarity is computed as the dot product between these vectors, and the model is run locally on the GPU.  
Both the sub-question answering and question reformulation LLMs use parameters from \autoref{tab:QD_LLM_parameters} with minor adjustments: the sub-question answering LLM employs a \texttt{repeat\_penalty} of 1.1 to mitigate repetitive output, while the reformulation module uses "?" as the stop token to restrict its output to a properly reformulated question.

\subsubsection{Answer Synthesis}
\paragraph{Overview}  
The final step synthesizes an answer to the original question using the generated reasoning chain, sub-questions, and sub-answers. This output, which includes the reasoning chain, provides transparency into the system's decision-making process.

\paragraph{Inputs and Outputs}  
Inputs comprise the main question, reasoning chain, sub-questions (reformulated if applicable), and sub-answers—all as strings. The output is a single natural language string that integrates both the final answer and the reasoning chain.

\paragraph{Techniques}  
A custom zero-shot prompt instructs the LLM to formulate the final answer from the provided context. The prompt template merges the main question, sub-questions, and sub-answers, and subsequently incorporates the reasoning chain into the final output. This straightforward zero-shot approach was preferred over ICL due to the simplicity of the final synthesis task compared to the more complex decomposition step.

\paragraph{Implementation Details}  
The LLM parameters mirror those in \autoref{tab:QD_LLM_parameters}, with the exception of \texttt{max\_tokens}, which is increased to 512 to accommodate the typically more complex final answers.

\section{Experiments}
\label{sec:experiments}
The goal of our experiments is check whether the usefulness of a KG in question answering and whether our approach, i.e., using an additional question decomposition module, results in a better performance.
For this purpose, we use a widely-used Knowledge Graph Question Answering (KGQA) benchmark called MetaQA \cite{Zhang}. In order to verify whether we achieved our objectives, we assess three baselines: a stand-alone LLM, an LLM with an LLM-based question-answering module, and an LLM with a KG (i.e., KAPING).
Eventually, the experimental results are presented and discussed.

\subsection{Dataset}
\label{sec:benchmark}
The MetaQA benchmark, introduced in 2017, addresses the need for KGQA benchmarks featuring multi-hop questions over large-scale KGs, extending the original WikiMovies benchmark with movie-domain questions of varying hop counts \cite{Zhang}.

Several factors motivated the selection of MetaQA for this research. First, its questions are categorized by hop count, enabling detailed analysis of multi-hop performance, a key area for improvement via question decomposition. Second, each question includes an entity label, avoiding the complexities of entity linking; many benchmarks, which focus on neural semantic parsing for SPARQL query generation, lack such labels \cite{KGQA_benchmarks}. Third, MetaQA’s simplicity and locally processable KG make it ideal for studies with limited resources, in contrast to highly complex KGs like Wikidata (over 130 GB, 1.57 billion triples, 12,175 relation types\footnote{\url{https://www.wikidata.org/wiki/Wikidata:Main_Page}}).

\paragraph{Data}  
MetaQA consists of three datasets (1-hop, 2-hop, and 3-hop), each split into train, validation, and test sets, and further divided into three components: vanilla, NTM text data, and audio data \cite{Zhang}. This research utilizes only the vanilla data, where the 1-hop dataset contains original WikiMovies questions and the 2-hop and 3-hop datasets are generated using predefined templates. Each dataset row includes a question, its associated entity, and answer entities.

\begin{figure}[h!]
    \centering
    \includegraphics[width=0.6\linewidth]{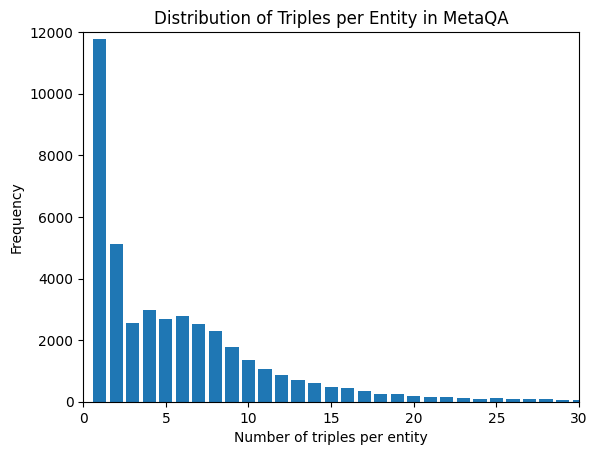}
    \caption{The distribution of degrees (triples per entity) in the MetaQA KG. (Note that the distribution is long-tailed, so the cut-off at the value of 30 is for the purpose of visualization.)}
    \label{fig:MetaQA_KG}
\end{figure}

\paragraph{Knowledge Graph}  
The MetaQA benchmark provides a KG as a text file with each row representing a triple. The KG comprises 43,234 entities and 9 relation types, with movie titles as subjects. \autoref{fig:MetaQA_KG} illustrates the degree distribution: most entities have few associated triples (median of 4), while the long-tailed distribution includes entities with up to 4431 triples.

\subsection{Experimental design}
In this study, we carry out two experiments:
\begin{enumerate}
    \item The goal of experiment 1 is to find out how the model parameters impact performance, in order to find a parameter configuration that leads to consistent performance over the different question types. The chosen parameter configuration can then be used to compare the system to baselines in the second experiment. 
    \item The main goal of the second experiment is to find out how different components of the system impact performance and overall behavior. This is achieved by comparing the performance of the system with specific baselines, which are essentially made up of combinations of system components.
\end{enumerate}

\subsubsection{Experiment 1: Model selection}  
Experiment 1 investigates the effect of model parameters on performance to determine a configuration that yields consistent results across different question types. The parameters under examination are the number of hops $N$ for candidate triple retrieval (tested with values 1, 2, 3) and the number of top triples $K$ selected for each sub-question (tested with values 10, 20, 30), consistent with values reported in the literature (\autoref{sec:related_work}).

For each MetaQA test dataset, 100 questions are sampled using a fixed seed, and the system is evaluated across all parameter combinations. This process is repeated with 10 different seeds (0–9) to capture performance variability, and all LLM components use the same seed for inference to ensure reproducibility.

Performance is measured using the Hit@1 metric, which checks if the generated answer exactly matches any of the label answer entities (after lowercasing and stripping). For example, if the label is "Brad Pitt" and the generated answer is "Pitt is the actor in question," the response is deemed incorrect. The final score for each dataset sample is the average Hit@1.

\subsubsection{Experiment 2: A Comparative Analysis with Baselines}
\label{sec:baselines}
Experiment 2 serves a purpose of assessing how individual system components influence overall performance by comparing the full system to three baselines:

\begin{enumerate}
    \item \textbf{LLM:} Uses only an LLM with a simple zero-shot prompt to directly answer the question.
    \item \textbf{LLM+QD:} Incorporates the question decomposition module to split questions and reformulate sub-questions before answering with the same zero-shot prompt as the LLM baseline.
    \item \textbf{LLM+KG:} Functions as the full system without the question decomposition component, which is equivalent to KAPING \cite{Baek2023} by employing candidate triple retrieval, top-$K$ triple selection, and the sub-question answering module.
\end{enumerate}

Both the full system and the LLM+KG baseline use the parameter configuration selected in \autoref{sec:exp1_results}. As in Experiment 1, 500 questions are sampled per MetaQA dataset using 8 different seeds (0–7) to ensure consistency. Performance is quantitatively evaluated using the Hit@1 metric to determine the impact of different components, and results are qualitatively analyzed for error insights and to assess accuracy, explainability, and generalizability as outlined in \autoref{sec:formal_problem}.

\subsection{Results and Discussion}

\subsubsection{Experiment 1: Quantitative analysis}
\label{sec:exp1_results}
\begin{figure}[h!]
    \centering
    \begin{subfigure}[b]{0.45\textwidth}
        \includegraphics[width=\textwidth]{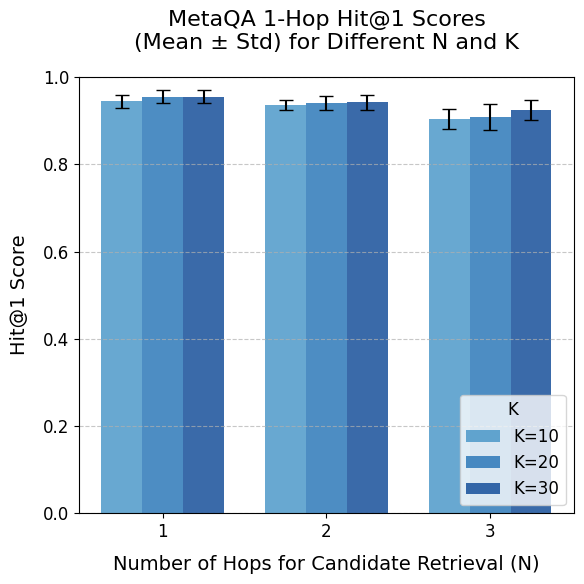}
    \end{subfigure}
    \hspace{0.05\textwidth} 
    \begin{subfigure}[b]{0.45\textwidth}
        \includegraphics[width=\textwidth]{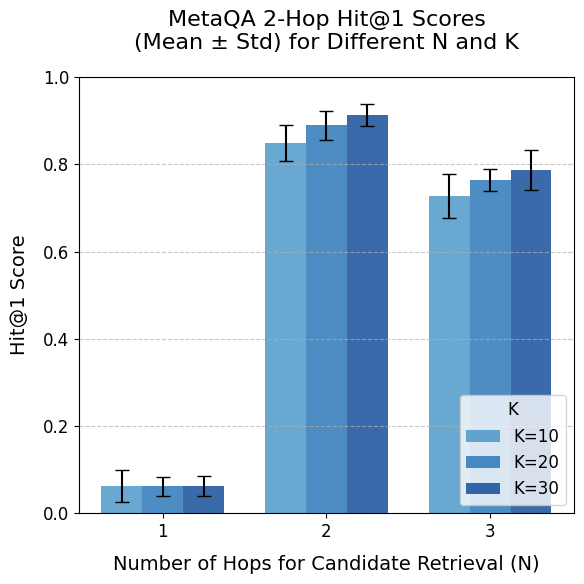}
    \end{subfigure}
    \vspace{0.5cm} 
    \begin{subfigure}[b]{0.45\textwidth}
        \includegraphics[width=\textwidth]{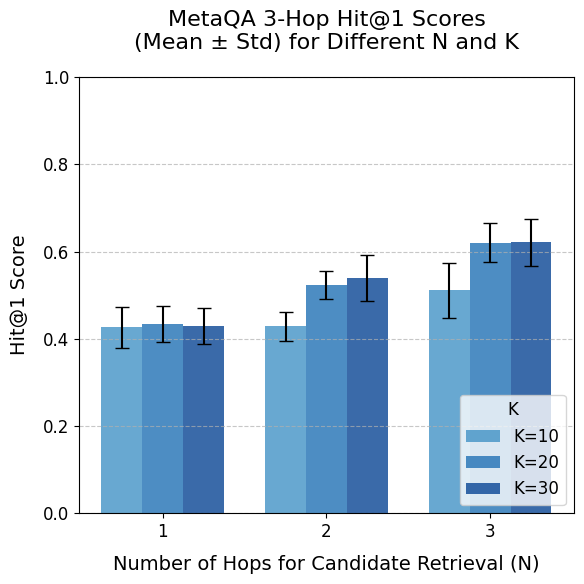}
    \end{subfigure}
    \caption{MetaQA performance results for experiment 1, over 10 samples of 100 questions for each of the three datasets. The bars show the mean Hit@1 for different parameter configurations; the error bars show the standard deviation.}
    \label{fig:results_exp1}
\end{figure}

The results of Experiment 1 (\autoref{fig:results_exp1}) indicate high overall performance that decreases with increasing question complexity, with standard deviations remaining low ($\leq 0.063$) across samples.

Performance is highest when the parameter $N$ equals the actual number of hops in the questions. As expected, for the 2-hop dataset, $N=1$ yields poor results; however, for the 3-hop dataset, performance with $N<3$ is unexpectedly high due to MetaQA’s question templates--for instance, some 3-hop questions (e.g., \textit{"Who are the directors of the films written by the writer of Blue Collar?"}) can be answered with $N=1$ triples. This represents a limitation of the MetaQA benchmark.

When holding the dataset and $N$ constant, increasing $K$ (the number of top triples selected) from 10 to 30 shows minimal effect on the 1-hop dataset, with slight improvements observed for the 2-hop and 3-hop datasets. Given that a higher $K$ is unlikely to reduce performance and is more likely to include the necessary triples, $K=30$ is chosen.

Considering the trade-offs across datasets, a balanced configuration is selected. Since $N=1$ is unacceptable for 2-hop questions and improved performance on 3-hop questions likely requires all candidate triples up to 3 hops, $N=3$ is deemed the best choice despite a minor reduction in 2-hop performance (0.787 $\pm$ 0.046). Consequently, the optimal parameter configuration for MetaQA is $N=3$ and $K=30$.

\subsubsection{Experiment 2: Quantitative analysis}
\label{sec:exp2}
\begin{figure}[h!]
    \centering
    \includegraphics[width=0.9\linewidth]{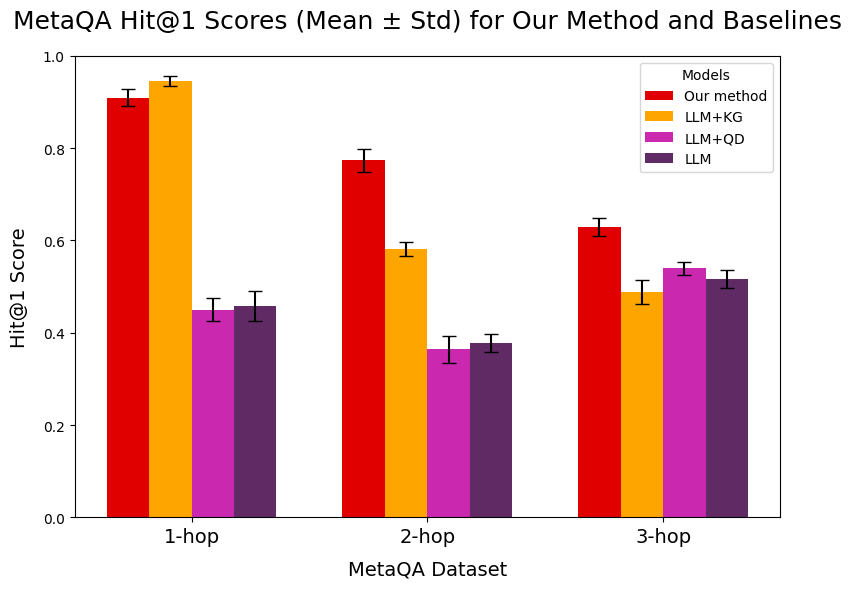}
    \caption{MetaQA performance results for experiment 2, over 8 samples of 500 questions for each of the three datasets. The bars show the mean Hit@1, and the error bars show the standard deviation. The results for both the system and the baselines are shown.}
    \label{fig:results_exp2_MetaQA}
\end{figure}

\autoref{fig:results_exp2_MetaQA} presents the performance results for Experiment 2 across 8 samples of 500 questions per MetaQA dataset. Our system significantly outperforms the baselines on 2-hop and 3-hop questions with minimal variance, while the LLM+KG baseline slightly outperforms on 1-hop questions. This is expected, as question decomposition adds unnecessary overhead for simple queries.


Comparing the baselines, the advantage of the KG retrieval module is most pronounced for 1-hop questions, but diminishes for 2-hop questions and disappears for 3-hop questions—likely because complex queries increase the difficulty of retrieving relevant triples. The integration of question decomposition in our system, however, maintains the benefits of KG retrieval for multi-hop questions while also enhancing answer explainability.

In summary, our system achieves improved performance on multi-hop questions with only a minor loss for 1-hop queries compared to the LLM+KG baseline. Although the relative and absolute advantage decreases as the number of hops increases, these quantitative results, combined with a forthcoming qualitative analysis (\autoref{sec:quali}), support the effectiveness of our approach.

\subsection{Qualitative Analysis}
\label{sec:quali}
This section examines the model outputs to identify recurring behaviors, strengths, and weaknesses, and to suggest directions for future improvements. Given the inherent limitations of a small, quantized LLM, our focus is on common patterns rather than isolated errors.

\begin{table}[h!]
\centering
\caption{The datasets that were analyzed for the qualitative analysis.}
\label{tab:qualitative_analysis}
\begin{tabular}{lllll}
\textbf{Dataset} & \textbf{Model} & \textbf{Experiment} & \textbf{Seed} & \textbf{Parameters} \\ \hline
MetaQA 1-hop     & KG-RAG        & 1                   & 0             & N=3, K=30           \\
MetaQA 2-hop     & KG-RAG         & 1                   & 0             & N=3, K=30           \\
MetaQA 3-hop     & KG-RAG         & 1                   & 0             & N=3, K=30           \\
\end{tabular}
\end{table}

\begin{figure}[h!]
    \centering
    \includegraphics[width=\linewidth]{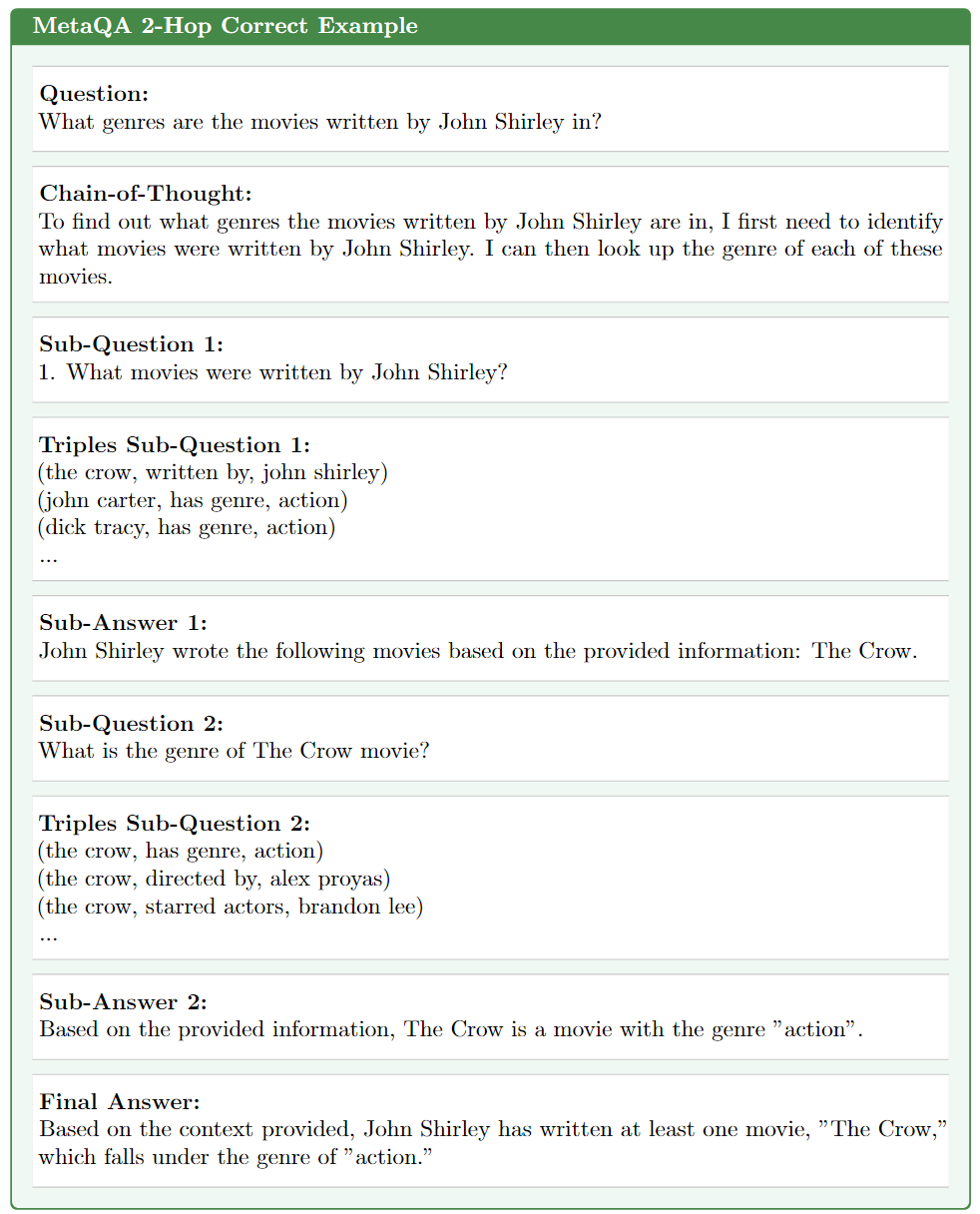}
    \caption{An example of the system's intermediate outputs, which lead to the final answer. The example was taken from the MetaQA 2-hop sample that was analyzed for the qualitative analysis.}
    \label{fig:output_example}
\end{figure}

\autoref{tab:qualitative_analysis} lists the generated outputs used in this analysis. First, we present an example output from the MetaQA 2-hop dataset (\autoref{fig:output_example}). This example demonstrates how the system correctly decomposes a question into coherent sub-questions, substitutes sub-answers appropriately, and synthesizes a final answer that accurately addresses the main question while maintaining a logical chain-of-thought.

\subsubsection{Question Decomposition}
\begin{figure}[h]
    \centering
    \includegraphics[width=1\linewidth]{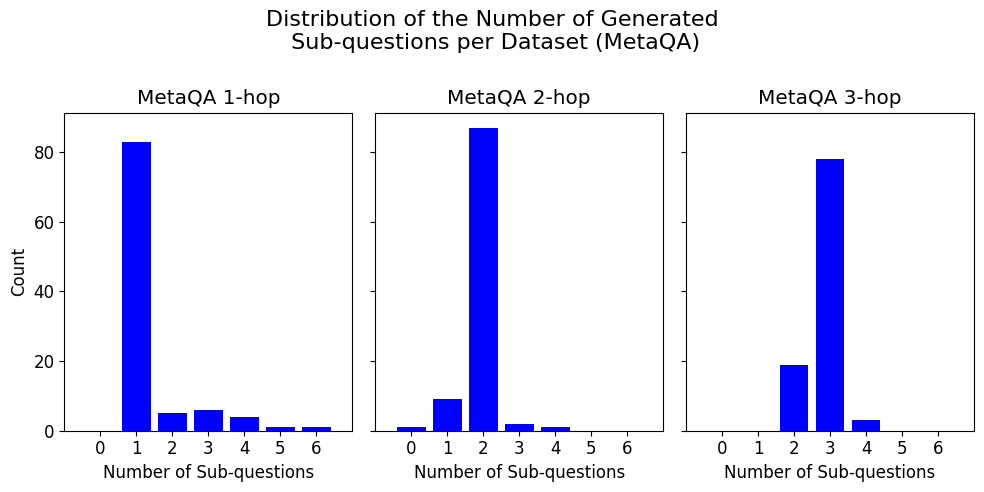}
    \caption{The distribution of the number of sub-questions that were generated, for each of the MetaQA samples that was analyzed for the qualitative analysis (see \autoref{tab:qualitative_analysis} for more details).}
    \label{fig:subq_distribution}
\end{figure}

By analyzing the distribution of the number of generated sub-questions per dataset (\autoref{fig:subq_distribution}), we observe that the model generally recognizes the appropriate complexity of MetaQA multi-hop questions. For 1-hop questions, the model typically avoids decomposition, though ambiguous queries (e.g. asking for a movie description) sometimes lead to unnecessary sub-questions. For 2-hop and 3-hop questions, the model usually generates the expected number of sub-questions, although there are occasional cases of under-decomposition.

\subsubsection{Qualitative Performance}
Overall, the system effectively distinguishes question complexity, but several systematic errors were identified:
\begin{itemize}
    \item \textbf{Over-decomposition:} In approximately 16\% of 1-hop cases, ambiguous questions lead to extra sub-questions, resulting in longer, sometimes overcomplicated answers.
    \item \textbf{Under-decomposition:} For 2-hop and 3-hop datasets, the system occasionally fails to generate enough sub-questions, sometimes producing only a 1-hop and a 2-hop question instead of the full decomposition.
    \item \textbf{Sub-answer Inconsistencies:} The LLM sometimes produces sub-answers that do not align with the provided triples, either by overlooking relevant data or by incorporating its own external knowledge.
    \item \textbf{Final Answer Synthesis:} While the final synthesis step generally succeeds, it occasionally yields overly long answers that may exceed token limits or include unwarranted information.
\end{itemize}

Despite these issues, the generated reasoning chains remain logical and coherent, allowing users to trace and verify the main answer. Many of the observed errors can be attributed to the limitations of the quantized LLM, and it is expected that a more sophisticated model or refined prompting strategies (potentially using ICL) could mitigate these problems.

In conclusion, while triple selection remains robust when question decomposition is successful, the identified issues in decomposition, sub-answer generation, and answer synthesis indicate clear avenues for future research and improvements.

\subsection{Discussion: Limitations}
Here, we outline key limitations of the carried out research, which subsequently allow us to formulate future work.

First, constrained computational resources forced the use of a quantized, relatively small LLM, significantly impacting absolute performance—despite potentially preserving relative improvements over baselines. These constraints also necessitated random sampling of test subsets rather than evaluating on full datasets.

Second, the MetaQA benchmark is relatively simple, with a narrow domain and exclusively multi-hop questions. As noted in \autoref{sec:exp1_results}, some 3-hop questions are answerable using only 1-hop triples, which may skew performance evaluations compared to more complex benchmarks.

Additionally, earlier experiments with another dataset, the Mintaka benchmark, revealed that the Hit@1 metric can be inaccurate, particularly for comparative questions where the system generates full answers instead of single answer entities. This limitation, highlighted in related work \cite{Baek2023, Wu2023, Wang2023}, underscores the need for more sophisticated evaluation methods. Recent research on automated evaluation of natural language outputs \cite{Guo2023} may offer promising alternatives.

In summary, key limitations include the restricted LLM size, the simplicity and flaws of the MetaQA benchmark, and the inadequacy of the Hit@1 metric for modern KGQA systems.

\section{Conclusion}
\label{sec:conclusion}

\subsection{Contributions}
Our study addressed two primary research questions. First, we investigated enhancing LLMs with knowledge graphs (KGs) without requiring any training. By leveraging the synergy between LLM reasoning and the structured knowledge in KGs, we identified a gap in creating KGQA models that are both generalizable (as in KAPING) and explainable (as in Keqing). To bridge this gap, we developed an improved, training-free version of KAPING.

Second, we explored methods to improve answer explainability using a KG-RAG system. Inspired by Keqing \cite{Wang2023} and the work of \cite{Radha2023}, we designed a question decomposition module that first generates a chain-of-thought (CoT) followed by coherent sub-questions. This approach not only improved performance on multi-hop questions but also provided transparent reasoning chains, thereby enhancing answer explainability. Overall, the proposed solution achieved higher answer accuracy (as measured by the Hit@1 metric) and improved transparency, though further validation is needed to confirm its generalizability across different domains, KGs, and question types.

\subsection{Future Work}
Future research should focus on deepening the investigation into application generalizability by employing benchmarks with KGs composed largely of natural language, ensuring the triple selection mechanism via text embeddings functions effectively. Given some limitations of the MetaQA benchmark, exploring alternative benchmarks with diverse question domains may yield more robust conclusions.

Improved evaluation methods are also necessary. Automated techniques—such as entity matching, coherence assessment of reasoning chains via LLM prompting, and verification of sub-answer validity—could offer more reliable metrics than the currently used Hit@1 \cite{Guo2023}.

Furthermore, employing more sophisticated LLMs and fine-tuning inference parameters could mitigate many of the systematic errors observed. Future work may also explore advanced, training-free triple retrieval methods or finetuning strategies for text embedding models, thereby enhancing performance and efficiency. Finally, addressing persistent research gaps in question entity identification and entity matching is crucial for real-world KGQA applications.

\bibliography{sn-bibliography}


\begin{thebibliography}{39}
\ifx \bisbn   \undefined \def \bisbn  #1{ISBN #1}\fi
\ifx \binits  \undefined \def \binits#1{#1}\fi
\ifx \bauthor  \undefined \def \bauthor#1{#1}\fi
\ifx \batitle  \undefined \def \batitle#1{#1}\fi
\ifx \bjtitle  \undefined \def \bjtitle#1{#1}\fi
\ifx \bvolume  \undefined \def \bvolume#1{\textbf{#1}}\fi
\ifx \byear  \undefined \def \byear#1{#1}\fi
\ifx \bissue  \undefined \def \bissue#1{#1}\fi
\ifx \bfpage  \undefined \def \bfpage#1{#1}\fi
\ifx \blpage  \undefined \def \blpage #1{#1}\fi
\ifx \burl  \undefined \def \burl#1{\textsf{#1}}\fi
\ifx \doiurl  \undefined \def \doiurl#1{\url{https://doi.org/#1}}\fi
\ifx \betal  \undefined \def \betal{\textit{et al.}}\fi
\ifx \binstitute  \undefined \def \binstitute#1{#1}\fi
\ifx \binstitutionaled  \undefined \def \binstitutionaled#1{#1}\fi
\ifx \bctitle  \undefined \def \bctitle#1{#1}\fi
\ifx \beditor  \undefined \def \beditor#1{#1}\fi
\ifx \bpublisher  \undefined \def \bpublisher#1{#1}\fi
\ifx \bbtitle  \undefined \def \bbtitle#1{#1}\fi
\ifx \bedition  \undefined \def \bedition#1{#1}\fi
\ifx \bseriesno  \undefined \def \bseriesno#1{#1}\fi
\ifx \blocation  \undefined \def \blocation#1{#1}\fi
\ifx \bsertitle  \undefined \def \bsertitle#1{#1}\fi
\ifx \bsnm \undefined \def \bsnm#1{#1}\fi
\ifx \bsuffix \undefined \def \bsuffix#1{#1}\fi
\ifx \bparticle \undefined \def \bparticle#1{#1}\fi
\ifx \barticle \undefined \def \barticle#1{#1}\fi
\bibcommenthead
\ifx \bconfdate \undefined \def \bconfdate #1{#1}\fi
\ifx \botherref \undefined \def \botherref #1{#1}\fi
\ifx \url \undefined \def \url#1{\textsf{#1}}\fi
\ifx \bchapter \undefined \def \bchapter#1{#1}\fi
\ifx \bbook \undefined \def \bbook#1{#1}\fi
\ifx \bcomment \undefined \def \bcomment#1{#1}\fi
\ifx \oauthor \undefined \def \oauthor#1{#1}\fi
\ifx \citeauthoryear \undefined \def \citeauthoryear#1{#1}\fi
\ifx \endbibitem  \undefined \def \endbibitem {}\fi
\ifx \bconflocation  \undefined \def \bconflocation#1{#1}\fi
\ifx \arxivurl  \undefined \def \arxivurl#1{\textsf{#1}}\fi
\csname PreBibitemsHook\endcsname

\bibitem[\protect\citeauthoryear{Brown et~al.}{2020}]{Brown2020}
\begin{botherref}
\oauthor{\bsnm{Brown}, \binits{T.B.}},
\oauthor{\bsnm{Mann}, \binits{B.}},
\oauthor{\bsnm{Ryder}, \binits{N.}},
\oauthor{\bsnm{Subbiah}, \binits{M.}},
\oauthor{\bsnm{Kaplan}, \binits{J.}},
\oauthor{\bsnm{Dhariwal}, \binits{P.}},
\oauthor{\bsnm{Neelakantan}, \binits{A.}},
\oauthor{\bsnm{Shyam}, \binits{P.}},
\oauthor{\bsnm{Sastry}, \binits{G.}},
\oauthor{\bsnm{Askell}, \binits{A.}},
\oauthor{\bsnm{Agarwal}, \binits{S.}},
\oauthor{\bsnm{Herbert-Voss}, \binits{A.}},
\oauthor{\bsnm{Krueger}, \binits{G.}},
\oauthor{\bsnm{Henighan}, \binits{T.}},
\oauthor{\bsnm{Child}, \binits{R.}},
\oauthor{\bsnm{Ramesh}, \binits{A.}},
\oauthor{\bsnm{Ziegler}, \binits{D.M.}},
\oauthor{\bsnm{Wu}, \binits{J.}},
\oauthor{\bsnm{Winter}, \binits{C.}},
\oauthor{\bsnm{Hesse}, \binits{C.}},
\oauthor{\bsnm{Chen}, \binits{M.}},
\oauthor{\bsnm{Sigler}, \binits{E.}},
\oauthor{\bsnm{Litwin}, \binits{M.}},
\oauthor{\bsnm{Gray}, \binits{S.}},
\oauthor{\bsnm{Chess}, \binits{B.}},
\oauthor{\bsnm{Clark}, \binits{J.}},
\oauthor{\bsnm{Berner}, \binits{C.}},
\oauthor{\bsnm{McCandlish}, \binits{S.}},
\oauthor{\bsnm{Radford}, \binits{A.}},
\oauthor{\bsnm{Sutskever}, \binits{I.}},
\oauthor{\bsnm{Amodei}, \binits{D.}}:
{Language models are few-shot learners}.
Advances in Neural Information Processing Systems
\textbf{2020-Decem}
(2020)
\end{botherref}
\endbibitem

\bibitem[\protect\citeauthoryear{Ji et~al.}{2023}]{Ji2023a}
\begin{botherref}
\oauthor{\bsnm{Ji}, \binits{Z.}},
\oauthor{\bsnm{Lee}, \binits{N.}},
\oauthor{\bsnm{Frieske}, \binits{R.}},
\oauthor{\bsnm{Yu}, \binits{T.}},
\oauthor{\bsnm{Su}, \binits{D.}},
\oauthor{\bsnm{Xu}, \binits{Y.}},
\oauthor{\bsnm{Ishii}, \binits{E.}},
\oauthor{\bsnm{Bang}, \binits{Y.J.}},
\oauthor{\bsnm{Madotto}, \binits{A.}},
\oauthor{\bsnm{Fung}, \binits{P.}}:
{Survey of Hallucination in Natural Language Generation}.
ACM Computing Surveys
\textbf{55}(12)
(2023)
\doiurl{10.1145/3571730}
\end{botherref}
\endbibitem

\bibitem[\protect\citeauthoryear{Pan et~al.}{2023}]{PanKalo2023}
\begin{botherref}
\oauthor{\bsnm{Pan}, \binits{J.Z.}},
\oauthor{\bsnm{Razniewski}, \binits{S.}},
\oauthor{\bsnm{Kalo}, \binits{J.-C.}},
\oauthor{\bsnm{Singhania}, \binits{S.}},
\oauthor{\bsnm{Chen}, \binits{J.}},
\oauthor{\bsnm{Dietze}, \binits{S.}},
\oauthor{\bsnm{Jabeen}, \binits{H.}},
\oauthor{\bsnm{Omeliyanenko}, \binits{J.}},
\oauthor{\bsnm{Zhang}, \binits{W.}},
\oauthor{\bsnm{Lissandrini}, \binits{M.}},
\oauthor{\bsnm{Biswas}, \binits{R.}},
\oauthor{\bsnm{Melo}, \binits{G.}},
\oauthor{\bsnm{Bonifati}, \binits{A.}},
\oauthor{\bsnm{Vakaj}, \binits{E.}},
\oauthor{\bsnm{Dragoni}, \binits{M.}},
\oauthor{\bsnm{Graux}, \binits{D.}}:
{Large Language Models and Knowledge Graphs: Opportunities and Challenges}
\textbf{000}(111),
1--30
(2023)
\end{botherref}
\endbibitem

\bibitem[\protect\citeauthoryear{Bender et~al.}{2021}]{Bender2021}
\begin{botherref}
\oauthor{\bsnm{Bender}, \binits{E.M.}},
\oauthor{\bsnm{Gebru}, \binits{T.}},
\oauthor{\bsnm{McMillan-Major}, \binits{A.}},
\oauthor{\bsnm{Shmitchell}, \binits{S.}}:
{On the dangers of stochastic parrots: Can language models be too big?}
FAccT 2021 - Proceedings of the 2021 ACM Conference on Fairness, Accountability, and Transparency,
610--623
(2021)
\doiurl{10.1145/3442188.3445922}
\end{botherref}
\endbibitem

\bibitem[\protect\citeauthoryear{Pan et~al.}{2023}]{Pan2023}
\begin{botherref}
\oauthor{\bsnm{Pan}, \binits{S.}},
\oauthor{\bsnm{Luo}, \binits{L.}},
\oauthor{\bsnm{Wang}, \binits{Y.}},
\oauthor{\bsnm{Chen}, \binits{C.}},
\oauthor{\bsnm{Wang}, \binits{J.}},
\oauthor{\bsnm{Wu}, \binits{X.}}:
{Unifying Large Language Models and Knowledge Graphs: A Roadmap}
\textbf{14}(8),
1--29
(2023)
\end{botherref}
\endbibitem

\bibitem[\protect\citeauthoryear{Yang et~al.}{2023}]{Yang2023}
\begin{botherref}
\oauthor{\bsnm{Yang}, \binits{L.}},
\oauthor{\bsnm{Chen}, \binits{H.}},
\oauthor{\bsnm{Li}, \binits{Z.}},
\oauthor{\bsnm{Ding}, \binits{X.}},
\oauthor{\bsnm{Wu}, \binits{X.}}:
{ChatGPT is not Enough: Enhancing Large Language Models with Knowledge Graphs for Fact-aware Language Modeling}
\textbf{14}(8),
1--20
(2023)
\end{botherref}
\endbibitem

\bibitem[\protect\citeauthoryear{Lewis et~al.}{2020}]{Lewis2020}
\begin{botherref}
\oauthor{\bsnm{Lewis}, \binits{P.}},
\oauthor{\bsnm{Perez}, \binits{E.}},
\oauthor{\bsnm{Piktus}, \binits{A.}},
\oauthor{\bsnm{Petroni}, \binits{F.}},
\oauthor{\bsnm{Karpukhin}, \binits{V.}},
\oauthor{\bsnm{Goyal}, \binits{N.}},
\oauthor{\bsnm{K{\"{u}}ttler}, \binits{H.}},
\oauthor{\bsnm{Lewis}, \binits{M.}},
\oauthor{\bsnm{Yih}, \binits{W.T.}},
\oauthor{\bsnm{Rockt{\"{a}}schel}, \binits{T.}},
\oauthor{\bsnm{Riedel}, \binits{S.}},
\oauthor{\bsnm{Kiela}, \binits{D.}}:
{Retrieval-augmented generation for knowledge-intensive NLP tasks}.
Advances in Neural Information Processing Systems
\textbf{2020-Decem}
(2020)
\end{botherref}
\endbibitem

\bibitem[\protect\citeauthoryear{Gao et~al.}{2023}]{Gao2024}
\begin{botherref}
\oauthor{\bsnm{Gao}, \binits{Y.}},
\oauthor{\bsnm{Xiong}, \binits{Y.}},
\oauthor{\bsnm{Gao}, \binits{X.}},
\oauthor{\bsnm{Jia}, \binits{K.}},
\oauthor{\bsnm{Pan}, \binits{J.}},
\oauthor{\bsnm{Bi}, \binits{Y.}},
\oauthor{\bsnm{Dai}, \binits{Y.}},
\oauthor{\bsnm{Sun}, \binits{J.}},
\oauthor{\bsnm{Wang}, \binits{M.}},
\oauthor{\bsnm{Wang}, \binits{H.}}:
{Retrieval-Augmented Generation for Large Language Models: A Survey}
(2023)
\end{botherref}
\endbibitem

\bibitem[\protect\citeauthoryear{Baek et~al.}{2023}]{Baek2023}
\begin{botherref}
\oauthor{\bsnm{Baek}, \binits{J.}},
\oauthor{\bsnm{Aji}, \binits{A.F.}},
\oauthor{\bsnm{Saffari}, \binits{A.}}:
{Knowledge-Augmented Language Model Prompting for Zero-Shot Knowledge Graph Question Answering}.
Proceedings of the Annual Meeting of the Association for Computational Linguistics,
70--98
(2023)
\doiurl{10.18653/v1/2023.nlrse-1.7}
\end{botherref}
\endbibitem

\bibitem[\protect\citeauthoryear{Wang et~al.}{2023}]{Wang2023}
\begin{botherref}
\oauthor{\bsnm{Wang}, \binits{C.}},
\oauthor{\bsnm{Xu}, \binits{Y.}},
\oauthor{\bsnm{Peng}, \binits{Z.}},
\oauthor{\bsnm{Zhang}, \binits{C.}},
\oauthor{\bsnm{Chen}, \binits{B.}},
\oauthor{\bsnm{Wang}, \binits{X.}},
\oauthor{\bsnm{Feng}, \binits{L.}},
\oauthor{\bsnm{An}, \binits{B.}}:
{keqing: knowledge-based question answering is a nature chain-of-thought mentor of LLM}
(2023)
\end{botherref}
\endbibitem

\bibitem[\protect\citeauthoryear{Wu et~al.}{2023}]{Wu2023}
\begin{botherref}
\oauthor{\bsnm{Wu}, \binits{Y.}},
\oauthor{\bsnm{Hu}, \binits{N.}},
\oauthor{\bsnm{Bi}, \binits{S.}},
\oauthor{\bsnm{Qi}, \binits{G.}},
\oauthor{\bsnm{Ren}, \binits{J.}},
\oauthor{\bsnm{Xie}, \binits{A.}},
\oauthor{\bsnm{Song}, \binits{W.}}:
{Retrieve-Rewrite-Answer: A KG-to-Text Enhanced LLMs Framework for Knowledge Graph Question Answering}
(2023)
\end{botherref}
\endbibitem

\bibitem[\protect\citeauthoryear{Arenas and Perez}{2013}]{arenas2013}
\begin{botherref}
\oauthor{\bsnm{Arenas}, \binits{M.}},
\oauthor{\bsnm{Perez}, \binits{J.}}:
{Querying Semantic Web Data with SPARQL}.
ACM
(2013)
\end{botherref}
\endbibitem

\bibitem[\protect\citeauthoryear{Vaswani et~al.}{2017}]{Vaswani2017}
\begin{botherref}
\oauthor{\bsnm{Vaswani}, \binits{A.}},
\oauthor{\bsnm{Shazeer}, \binits{N.}},
\oauthor{\bsnm{Parmar}, \binits{N.}},
\oauthor{\bsnm{Uszkoreit}, \binits{J.}},
\oauthor{\bsnm{Jones}, \binits{L.}},
\oauthor{\bsnm{Gomez}, \binits{A.N.}},
\oauthor{\bsnm{Kaiser}, \binits{L.}},
\oauthor{\bsnm{Polosukhin}, \binits{I.}}:
Attention is all you need.
Advances in neural information processing systems
\textbf{30}
(2017)
\end{botherref}
\endbibitem

\bibitem[\protect\citeauthoryear{Kaplan et~al.}{2020}]{kaplan2020}
\begin{botherref}
\oauthor{\bsnm{Kaplan}, \binits{J.}},
\oauthor{\bsnm{McCandlish}, \binits{S.}},
\oauthor{\bsnm{Henighan}, \binits{T.}},
\oauthor{\bsnm{Brown}, \binits{T.B.}},
\oauthor{\bsnm{Chess}, \binits{B.}},
\oauthor{\bsnm{Child}, \binits{R.}},
\oauthor{\bsnm{Gray}, \binits{S.}},
\oauthor{\bsnm{Radford}, \binits{A.}},
\oauthor{\bsnm{Wu}, \binits{J.}},
\oauthor{\bsnm{Amodei}, \binits{D.}}:
{Scaling Laws for Neural Language Models}
(2020)
\end{botherref}
\endbibitem

\bibitem[\protect\citeauthoryear{Zhao et~al.}{2023}]{Zhao2023}
\begin{botherref}
\oauthor{\bsnm{Zhao}, \binits{W.X.}},
\oauthor{\bsnm{Zhou}, \binits{K.}},
\oauthor{\bsnm{Li}, \binits{J.}},
\oauthor{\bsnm{Tang}, \binits{T.}},
\oauthor{\bsnm{Wang}, \binits{X.}},
\oauthor{\bsnm{Hou}, \binits{Y.}},
\oauthor{\bsnm{Min}, \binits{Y.}},
\oauthor{\bsnm{Zhang}, \binits{B.}},
\oauthor{\bsnm{Zhang}, \binits{J.}},
\oauthor{\bsnm{Dong}, \binits{Z.}},
\oauthor{\bsnm{Du}, \binits{Y.}},
\oauthor{\bsnm{Yang}, \binits{C.}},
\oauthor{\bsnm{Chen}, \binits{Y.}},
\oauthor{\bsnm{Chen}, \binits{Z.}},
\oauthor{\bsnm{Jiang}, \binits{J.}},
\oauthor{\bsnm{Ren}, \binits{R.}},
\oauthor{\bsnm{Li}, \binits{Y.}},
\oauthor{\bsnm{Tang}, \binits{X.}},
\oauthor{\bsnm{Liu}, \binits{Z.}},
\oauthor{\bsnm{Liu}, \binits{P.}},
\oauthor{\bsnm{Nie}, \binits{J.-Y.}},
\oauthor{\bsnm{Wen}, \binits{J.-R.}}:
{A Survey of Large Language Models},
1--97
(2023)
\end{botherref}
\endbibitem

\bibitem[\protect\citeauthoryear{Chen et~al.}{}]{chen2023}
\begin{botherref}
\oauthor{\bsnm{Chen}, \binits{J.}},
\oauthor{\bsnm{Chen}, \binits{L.}},
\oauthor{\bsnm{Zhu}, \binits{C.}},
\oauthor{\bsnm{Zhou}, \binits{T.}}:
{How Many Demonstrations Do You Need for In-context Learning?}
Technical report
\end{botherref}
\endbibitem

\bibitem[\protect\citeauthoryear{Wei et~al.}{2022}]{Wei2023}
\begin{botherref}
\oauthor{\bsnm{Wei}, \binits{J.}},
\oauthor{\bsnm{Wang}, \binits{X.}},
\oauthor{\bsnm{Schuurmans}, \binits{D.}},
\oauthor{\bsnm{Bosma}, \binits{M.}},
\oauthor{\bsnm{Ichter}, \binits{B.}},
\oauthor{\bsnm{Xia}, \binits{F.}},
\oauthor{\bsnm{Chi}, \binits{E.}},
\oauthor{\bsnm{Le}, \binits{Q.}},
\oauthor{\bsnm{Zhou}, \binits{D.}}:
{Chain-of-Thought Prompting Elicits Reasoning in Large Language Models}
(2022)
\end{botherref}
\endbibitem

\bibitem[\protect\citeauthoryear{Sen et~al.}{2022}]{Sen2022}
\begin{barticle}
\bauthor{\bsnm{Sen}, \binits{P.}},
\bauthor{\bsnm{Aji}, \binits{A.F.}},
\bauthor{\bsnm{Saffari}, \binits{A.}}:
\batitle{{Mintaka: A Complex, Natural, and Multilingual Dataset for End-to-End Question Answering}}.
\bjtitle{Proceedings - International Conference on Computational Linguistics, COLING}
\bvolume{29}(\bissue{1}),
\bfpage{1604}--\blpage{1619}
(\byear{2022})
\end{barticle}
\endbibitem

\bibitem[\protect\citeauthoryear{Yih et~al.}{2016}]{Yih2016}
\begin{botherref}
\oauthor{\bsnm{Yih}, \binits{W.T.}},
\oauthor{\bsnm{Richardson}, \binits{M.}},
\oauthor{\bsnm{Meek}, \binits{C.}},
\oauthor{\bsnm{Chang}, \binits{M.W.}},
\oauthor{\bsnm{Suh}, \binits{J.}}:
{The value of semantic parse labeling for knowledge base question answering}.
54th Annual Meeting of the Association for Computational Linguistics, ACL 2016 - Short Papers,
201--206
(2016)
\doiurl{10.18653/v1/p16-2033}
\end{botherref}
\endbibitem

\bibitem[\protect\citeauthoryear{Zhang et~al.}{}]{Zhang}
\begin{botherref}
\oauthor{\bsnm{Zhang}, \binits{Y.}},
\oauthor{\bsnm{Dai}, \binits{H.}},
\oauthor{\bsnm{Kozareva}, \binits{Z.}},
\oauthor{\bsnm{Smola}, \binits{A.J.}},
\oauthor{\bsnm{Song}, \binits{L.}}:
{Variational Reasoning for Question Answering with Knowledge Graph},
1--22
\end{botherref}
\endbibitem

\bibitem[\protect\citeauthoryear{Oliya et~al.}{}]{Oliya2021}
\begin{botherref}
\oauthor{\bsnm{Oliya}, \binits{A.}},
\oauthor{\bsnm{Saffari}, \binits{A.}},
\oauthor{\bsnm{Sen}, \binits{P.}},
\oauthor{\bsnm{Ayoola}, \binits{T.}}:
{End-to-End Entity Resolution and Question Answering Using Differentiable Knowledge Graphs}.
Technical report
\end{botherref}
\endbibitem

\bibitem[\protect\citeauthoryear{Sen et~al.}{2023}]{Sen2023}
\begin{botherref}
\oauthor{\bsnm{Sen}, \binits{P.}},
\oauthor{\bsnm{Mavadia}, \binits{S.}},
\oauthor{\bsnm{Saffari}, \binits{A.}}:
{Knowledge Graph-augmented Language Models for Complex Question Answering}.
Proceedings of the Annual Meeting of the Association for Computational Linguistics,
1--8
(2023)
\doiurl{10.18653/v1/2023.nlrse-1.1}
\end{botherref}
\endbibitem

\bibitem[\protect\citeauthoryear{Gu et~al.}{2022}]{Gu2022}
\begin{botherref}
\oauthor{\bsnm{Gu}, \binits{Y.}},
\oauthor{\bsnm{Pahuja}, \binits{V.}},
\oauthor{\bsnm{Cheng}, \binits{G.}},
\oauthor{\bsnm{Su}, \binits{Y.}}:
{Knowledge Base Question Answering: A Semantic Parsing Perspective}
(2022)
\end{botherref}
\endbibitem

\bibitem[\protect\citeauthoryear{Sanh et~al.}{2022}]{Sanh2022}
\begin{botherref}
\oauthor{\bsnm{Sanh}, \binits{V.}},
\oauthor{\bsnm{Webson}, \binits{A.}},
\oauthor{\bsnm{Raffel}, \binits{C.}},
\oauthor{\bsnm{Bach}, \binits{S.H.}},
\oauthor{\bsnm{Sutawika}, \binits{L.}},
\oauthor{\bsnm{Alyafeai}, \binits{Z.}},
\oauthor{\bsnm{Chaffin}, \binits{A.}},
\oauthor{\bsnm{Stiegler}, \binits{A.}},
\oauthor{\bsnm{Le~Scao}, \binits{T.}},
\oauthor{\bsnm{Raja}, \binits{A.}},
\oauthor{\bsnm{Dey}, \binits{M.}},
\oauthor{\bsnm{Bari}, \binits{M.S.}},
\oauthor{\bsnm{Xu}, \binits{C.}},
\oauthor{\bsnm{Thakker}, \binits{U.}},
\oauthor{\bsnm{Sharma}, \binits{S.}},
\oauthor{\bsnm{Szczechla}, \binits{E.}},
\oauthor{\bsnm{Kim}, \binits{T.}},
\oauthor{\bsnm{Chhablani}, \binits{G.}},
\oauthor{\bsnm{Nayak}, \binits{N.V.}},
\oauthor{\bsnm{Datta}, \binits{D.}},
\oauthor{\bsnm{Chang}, \binits{J.}},
\oauthor{\bsnm{Jiang}, \binits{M.T.J.}},
\oauthor{\bsnm{Wang}, \binits{H.}},
\oauthor{\bsnm{Manica}, \binits{M.}},
\oauthor{\bsnm{Shen}, \binits{S.}},
\oauthor{\bsnm{Yong}, \binits{Z.X.}},
\oauthor{\bsnm{Pandey}, \binits{H.}},
\oauthor{\bsnm{McKenna}, \binits{M.}},
\oauthor{\bsnm{Bawden}, \binits{R.}},
\oauthor{\bsnm{Wang}, \binits{T.}},
\oauthor{\bsnm{Neeraj}, \binits{T.}},
\oauthor{\bsnm{Rozen}, \binits{J.}},
\oauthor{\bsnm{Sharma}, \binits{A.}},
\oauthor{\bsnm{Santilli}, \binits{A.}},
\oauthor{\bsnm{Fevry}, \binits{T.}},
\oauthor{\bsnm{Fries}, \binits{J.A.}},
\oauthor{\bsnm{Teehan}, \binits{R.}},
\oauthor{\bsnm{Bers}, \binits{T.}},
\oauthor{\bsnm{Biderman}, \binits{S.}},
\oauthor{\bsnm{Gao}, \binits{L.}},
\oauthor{\bsnm{Wolf}, \binits{T.}},
\oauthor{\bsnm{Rush}, \binits{A.M.}}:
{Multitask Prompted Training Enables Zero-Shot Task Generalization}.
ICLR 2022 - 10th International Conference on Learning Representations
(2022)
\end{botherref}
\endbibitem

\bibitem[\protect\citeauthoryear{Raffel et~al.}{2020}]{Raffel2020}
\begin{barticle}
\bauthor{\bsnm{Raffel}, \binits{C.}},
\bauthor{\bsnm{Shazeer}, \binits{N.}},
\bauthor{\bsnm{Roberts}, \binits{A.}},
\bauthor{\bsnm{Lee}, \binits{K.}},
\bauthor{\bsnm{Narang}, \binits{S.}},
\bauthor{\bsnm{Matena}, \binits{M.}},
\bauthor{\bsnm{Zhou}, \binits{Y.}},
\bauthor{\bsnm{Li}, \binits{W.}},
\bauthor{\bsnm{Liu}, \binits{P.J.}}:
\batitle{{Exploring the limits of transfer learning with a unified text-to-text transformer}}.
\bjtitle{Journal of Machine Learning Research}
\bvolume{21},
\bfpage{1}--\blpage{67}
(\byear{2020})
\end{barticle}
\endbibitem

\bibitem[\protect\citeauthoryear{Chung et~al.}{2022}]{Chung2022}
\begin{botherref}
\oauthor{\bsnm{Chung}, \binits{H.W.}},
\oauthor{\bsnm{Hou}, \binits{L.}},
\oauthor{\bsnm{Longpre}, \binits{S.}},
\oauthor{\bsnm{Zoph}, \binits{B.}},
\oauthor{\bsnm{Tay}, \binits{Y.}},
\oauthor{\bsnm{Fedus}, \binits{W.}},
\oauthor{\bsnm{Li}, \binits{Y.}},
\oauthor{\bsnm{Wang}, \binits{X.}},
\oauthor{\bsnm{Dehghani}, \binits{M.}},
\oauthor{\bsnm{Brahma}, \binits{S.}},
\oauthor{\bsnm{Webson}, \binits{A.}},
\oauthor{\bsnm{Gu}, \binits{S.S.}},
\oauthor{\bsnm{Dai}, \binits{Z.}},
\oauthor{\bsnm{Suzgun}, \binits{M.}},
\oauthor{\bsnm{Chen}, \binits{X.}},
\oauthor{\bsnm{Chowdhery}, \binits{A.}},
\oauthor{\bsnm{Castro-Ros}, \binits{A.}},
\oauthor{\bsnm{Pellat}, \binits{M.}},
\oauthor{\bsnm{Robinson}, \binits{K.}},
\oauthor{\bsnm{Valter}, \binits{D.}},
\oauthor{\bsnm{Narang}, \binits{S.}},
\oauthor{\bsnm{Mishra}, \binits{G.}},
\oauthor{\bsnm{Yu}, \binits{A.}},
\oauthor{\bsnm{Zhao}, \binits{V.}},
\oauthor{\bsnm{Huang}, \binits{Y.}},
\oauthor{\bsnm{Dai}, \binits{A.}},
\oauthor{\bsnm{Yu}, \binits{H.}},
\oauthor{\bsnm{Petrov}, \binits{S.}},
\oauthor{\bsnm{Chi}, \binits{E.H.}},
\oauthor{\bsnm{Dean}, \binits{J.}},
\oauthor{\bsnm{Devlin}, \binits{J.}},
\oauthor{\bsnm{Roberts}, \binits{A.}},
\oauthor{\bsnm{Zhou}, \binits{D.}},
\oauthor{\bsnm{Le}, \binits{Q.V.}},
\oauthor{\bsnm{Wei}, \binits{J.}}:
{Scaling Instruction-Finetuned Language Models},
1--54
(2022)
\end{botherref}
\endbibitem

\bibitem[\protect\citeauthoryear{Zhang et~al.}{2022}]{Zhang2022}
\begin{botherref}
\oauthor{\bsnm{Zhang}, \binits{S.}},
\oauthor{\bsnm{Roller}, \binits{S.}},
\oauthor{\bsnm{Goyal}, \binits{N.}},
\oauthor{\bsnm{Artetxe}, \binits{M.}},
\oauthor{\bsnm{Chen}, \binits{M.}},
\oauthor{\bsnm{Chen}, \binits{S.}},
\oauthor{\bsnm{Dewan}, \binits{C.}},
\oauthor{\bsnm{Diab}, \binits{M.}},
\oauthor{\bsnm{Li}, \binits{X.}},
\oauthor{\bsnm{Lin}, \binits{X.V.}},
\oauthor{\bsnm{Mihaylov}, \binits{T.}},
\oauthor{\bsnm{Ott}, \binits{M.}},
\oauthor{\bsnm{Shleifer}, \binits{S.}},
\oauthor{\bsnm{Shuster}, \binits{K.}},
\oauthor{\bsnm{Simig}, \binits{D.}},
\oauthor{\bsnm{Koura}, \binits{P.S.}},
\oauthor{\bsnm{Sridhar}, \binits{A.}},
\oauthor{\bsnm{Wang}, \binits{T.}},
\oauthor{\bsnm{Zettlemoyer}, \binits{L.}}:
{OPT: Open Pre-trained Transformer Language Models}
(2022)
\end{botherref}
\endbibitem

\bibitem[\protect\citeauthoryear{Fitzgerald et~al.}{2022}]{Fitzgerald2022}
\begin{botherref}
\oauthor{\bsnm{Fitzgerald}, \binits{J.}},
\oauthor{\bsnm{Ananthakrishnan}, \binits{S.}},
\oauthor{\bsnm{Arkoudas}, \binits{K.}},
\oauthor{\bsnm{Bernardi}, \binits{D.}},
\oauthor{\bsnm{Bhagia}, \binits{A.}},
\oauthor{\bsnm{Delli~Bovi}, \binits{C.}},
\oauthor{\bsnm{Cao}, \binits{J.}},
\oauthor{\bsnm{Chada}, \binits{R.}},
\oauthor{\bsnm{Chauhan}, \binits{A.}},
\oauthor{\bsnm{Chen}, \binits{L.}},
\oauthor{\bsnm{Dwarakanath}, \binits{A.}},
\oauthor{\bsnm{Dwivedi}, \binits{S.}},
\oauthor{\bsnm{Gojayev}, \binits{T.}},
\oauthor{\bsnm{Gopalakrishnan}, \binits{K.}},
\oauthor{\bsnm{Gueudre}, \binits{T.}},
\oauthor{\bsnm{Hakkani-Tur}, \binits{D.}},
\oauthor{\bsnm{Hamza}, \binits{W.}},
\oauthor{\bsnm{Hueser}, \binits{J.J.}},
\oauthor{\bsnm{Jose}, \binits{K.M.}},
\oauthor{\bsnm{Khan}, \binits{H.}},
\oauthor{\bsnm{Liu}, \binits{B.}},
\oauthor{\bsnm{Lu}, \binits{J.}},
\oauthor{\bsnm{Manzotti}, \binits{A.}},
\oauthor{\bsnm{Natarajan}, \binits{P.}},
\oauthor{\bsnm{Owczarzak}, \binits{K.}},
\oauthor{\bsnm{Oz}, \binits{G.}},
\oauthor{\bsnm{Palumbo}, \binits{E.}},
\oauthor{\bsnm{Peris}, \binits{C.}},
\oauthor{\bsnm{Prakash}, \binits{C.S.}},
\oauthor{\bsnm{Rawls}, \binits{S.}},
\oauthor{\bsnm{Rosenbaum}, \binits{A.}},
\oauthor{\bsnm{Shenoy}, \binits{A.}},
\oauthor{\bsnm{Soltan}, \binits{S.}},
\oauthor{\bsnm{Sridhar}, \binits{M.H.}},
\oauthor{\bsnm{Tan}, \binits{L.}},
\oauthor{\bsnm{Triefenbach}, \binits{F.}},
\oauthor{\bsnm{Wei}, \binits{P.}},
\oauthor{\bsnm{Yu}, \binits{H.}},
\oauthor{\bsnm{Zheng}, \binits{S.}},
\oauthor{\bsnm{Tur}, \binits{G.}},
\oauthor{\bsnm{Natarajan}, \binits{P.}}:
{Alexa Teacher Model: Pretraining and Distilling Multi-Billion-Parameter Encoders for Natural Language Understanding Systems}.
Proceedings of the ACM SIGKDD International Conference on Knowledge Discovery and Data Mining,
2893--2902
(2022)
\doiurl{10.1145/3534678.3539173}
\end{botherref}
\endbibitem

\bibitem[\protect\citeauthoryear{Touvron et~al.}{2023}]{Touvron2023}
\begin{botherref}
\oauthor{\bsnm{Touvron}, \binits{H.}},
\oauthor{\bsnm{Martin}, \binits{L.}},
\oauthor{\bsnm{Stone}, \binits{K.}},
\oauthor{\bsnm{Albert}, \binits{P.}},
\oauthor{\bsnm{Almahairi}, \binits{A.}},
\oauthor{\bsnm{Babaei}, \binits{Y.}},
\oauthor{\bsnm{Bashlykov}, \binits{N.}},
\oauthor{\bsnm{Batra}, \binits{S.}},
\oauthor{\bsnm{Bhargava}, \binits{P.}},
\oauthor{\bsnm{Bhosale}, \binits{S.}},
\oauthor{\bsnm{Bikel}, \binits{D.}},
\oauthor{\bsnm{Blecher}, \binits{L.}},
\oauthor{\bsnm{Ferrer}, \binits{C.C.}},
\oauthor{\bsnm{Chen}, \binits{M.}},
\oauthor{\bsnm{Cucurull}, \binits{G.}},
\oauthor{\bsnm{Esiobu}, \binits{D.}},
\oauthor{\bsnm{Fernandes}, \binits{J.}},
\oauthor{\bsnm{Fu}, \binits{J.}},
\oauthor{\bsnm{Fu}, \binits{W.}},
\oauthor{\bsnm{Fuller}, \binits{B.}},
\oauthor{\bsnm{Gao}, \binits{C.}},
\oauthor{\bsnm{Goswami}, \binits{V.}},
\oauthor{\bsnm{Goyal}, \binits{N.}},
\oauthor{\bsnm{Hartshorn}, \binits{A.}},
\oauthor{\bsnm{Hosseini}, \binits{S.}},
\oauthor{\bsnm{Hou}, \binits{R.}},
\oauthor{\bsnm{Inan}, \binits{H.}},
\oauthor{\bsnm{Kardas}, \binits{M.}},
\oauthor{\bsnm{Kerkez}, \binits{V.}},
\oauthor{\bsnm{Khabsa}, \binits{M.}},
\oauthor{\bsnm{Kloumann}, \binits{I.}},
\oauthor{\bsnm{Korenev}, \binits{A.}},
\oauthor{\bsnm{Koura}, \binits{P.S.}},
\oauthor{\bsnm{Lachaux}, \binits{M.-A.}},
\oauthor{\bsnm{Lavril}, \binits{T.}},
\oauthor{\bsnm{Lee}, \binits{J.}},
\oauthor{\bsnm{Liskovich}, \binits{D.}},
\oauthor{\bsnm{Lu}, \binits{Y.}},
\oauthor{\bsnm{Mao}, \binits{Y.}},
\oauthor{\bsnm{Martinet}, \binits{X.}},
\oauthor{\bsnm{Mihaylov}, \binits{T.}},
\oauthor{\bsnm{Mishra}, \binits{P.}},
\oauthor{\bsnm{Molybog}, \binits{I.}},
\oauthor{\bsnm{Nie}, \binits{Y.}},
\oauthor{\bsnm{Poulton}, \binits{A.}},
\oauthor{\bsnm{Reizenstein}, \binits{J.}},
\oauthor{\bsnm{Rungta}, \binits{R.}},
\oauthor{\bsnm{Saladi}, \binits{K.}},
\oauthor{\bsnm{Schelten}, \binits{A.}},
\oauthor{\bsnm{Silva}, \binits{R.}},
\oauthor{\bsnm{Smith}, \binits{E.M.}},
\oauthor{\bsnm{Subramanian}, \binits{R.}},
\oauthor{\bsnm{Tan}, \binits{X.E.}},
\oauthor{\bsnm{Tang}, \binits{B.}},
\oauthor{\bsnm{Taylor}, \binits{R.}},
\oauthor{\bsnm{Williams}, \binits{A.}},
\oauthor{\bsnm{Kuan}, \binits{J.X.}},
\oauthor{\bsnm{Xu}, \binits{P.}},
\oauthor{\bsnm{Yan}, \binits{Z.}},
\oauthor{\bsnm{Zarov}, \binits{I.}},
\oauthor{\bsnm{Zhang}, \binits{Y.}},
\oauthor{\bsnm{Fan}, \binits{A.}},
\oauthor{\bsnm{Kambadur}, \binits{M.}},
\oauthor{\bsnm{Narang}, \binits{S.}},
\oauthor{\bsnm{Rodriguez}, \binits{A.}},
\oauthor{\bsnm{Stojnic}, \binits{R.}},
\oauthor{\bsnm{Edunov}, \binits{S.}},
\oauthor{\bsnm{Scialom}, \binits{T.}}:
{Llama 2: Open Foundation and Fine-Tuned Chat Models}
(2023)
\end{botherref}
\endbibitem

\bibitem[\protect\citeauthoryear{Berant}{2013}]{Berant2013}
\begin{botherref}
\oauthor{\bsnm{Berant}, \binits{J.}}:
{Semantic Parsing on Freebase from Question-Answer Pairs}
(October),
1533--1544
(2013)
\end{botherref}
\endbibitem

\bibitem[\protect\citeauthoryear{Talmor and Berant}{2013}]{Talmor2013}
\begin{botherref}
\oauthor{\bsnm{Talmor}, \binits{A.}},
\oauthor{\bsnm{Berant}, \binits{J.}}:
{The Web as a Knowledge-base for Answering Complex Questions}
(2013)
\end{botherref}
\endbibitem

\bibitem[\protect\citeauthoryear{Dubey et~al.}{}]{Dubey}
\begin{botherref}
\oauthor{\bsnm{Dubey}, \binits{M.}},
\oauthor{\bsnm{Banerjee}, \binits{D.}},
\oauthor{\bsnm{Abdelkawi}, \binits{A.}}:
{LC-QuAD 2 . 0 : A Large Dataset for Complex Question Answering over Wikidata and DBpedia}
\end{botherref}
\endbibitem

\bibitem[\protect\citeauthoryear{Pedersen et~al.}{}]{Pedersen2004}
\begin{botherref}
\oauthor{\bsnm{Pedersen}, \binits{T.}},
\oauthor{\bsnm{Patwardhan}, \binits{S.}},
\oauthor{\bsnm{Michelizzi}, \binits{J.}}:
{WordNet::Similarity-Measuring the Relatedness of Concepts}.
Technical report.
\url{http://search.cpan.org/dist/WordNet-Similarityhttp://wn-similarity.sourceforge.net}
\end{botherref}
\endbibitem

\bibitem[\protect\citeauthoryear{Hu et~al.}{2021}]{Shen2021}
\begin{botherref}
\oauthor{\bsnm{Hu}, \binits{E.J.}},
\oauthor{\bsnm{Shen}, \binits{Y.}},
\oauthor{\bsnm{Wallis}, \binits{P.}},
\oauthor{\bsnm{Allen-Zhu}, \binits{Z.}},
\oauthor{\bsnm{Li}, \binits{Y.}},
\oauthor{\bsnm{Wang}, \binits{S.}},
\oauthor{\bsnm{Wang}, \binits{L.}},
\oauthor{\bsnm{Chen}, \binits{W.}}:
{LoRA: Low-Rank Adaptation of Large Language Models}
(2021)
\end{botherref}
\endbibitem

\bibitem[\protect\citeauthoryear{Radhakrishnan et~al.}{2023}]{Radha2023}
\begin{botherref}
\oauthor{\bsnm{Radhakrishnan}, \binits{A.}},
\oauthor{\bsnm{Nguyen}, \binits{K.}},
\oauthor{\bsnm{Chen}, \binits{A.}},
\oauthor{\bsnm{Chen}, \binits{C.}},
\oauthor{\bsnm{Denison}, \binits{C.}},
\oauthor{\bsnm{Hernandez}, \binits{D.}},
\oauthor{\bsnm{Durmus}, \binits{E.}},
\oauthor{\bsnm{Hubinger}, \binits{E.}},
\oauthor{\bsnm{Kernion}, \binits{J.}},
\oauthor{\bsnm{Luko{\v{s}}i{\={u}}t{\.{e}}}, \binits{K.}},
\oauthor{\bsnm{Cheng}, \binits{N.}},
\oauthor{\bsnm{Joseph}, \binits{N.}},
\oauthor{\bsnm{Schiefer}, \binits{N.}},
\oauthor{\bsnm{Rausch}, \binits{O.}},
\oauthor{\bsnm{McCandlish}, \binits{S.}},
\oauthor{\bsnm{Showk}, \binits{S.E.}},
\oauthor{\bsnm{Lanham}, \binits{T.}},
\oauthor{\bsnm{Maxwell}, \binits{T.}},
\oauthor{\bsnm{Chandrasekaran}, \binits{V.}},
\oauthor{\bsnm{Hatfield-Dodds}, \binits{Z.}},
\oauthor{\bsnm{Kaplan}, \binits{J.}},
\oauthor{\bsnm{Brauner}, \binits{J.}},
\oauthor{\bsnm{Bowman}, \binits{S.R.}},
\oauthor{\bsnm{Perez}, \binits{E.}}:
{Question Decomposition Improves the Faithfulness of Model-Generated Reasoning}
(2023)
\end{botherref}
\endbibitem

\bibitem[\protect\citeauthoryear{Jiang et~al.}{2023}]{Jiang2023}
\begin{botherref}
\oauthor{\bsnm{Jiang}, \binits{A.Q.}},
\oauthor{\bsnm{Sablayrolles}, \binits{A.}},
\oauthor{\bsnm{Mensch}, \binits{A.}},
\oauthor{\bsnm{Bamford}, \binits{C.}},
\oauthor{\bsnm{Chaplot}, \binits{D.S.}},
\oauthor{\bsnm{Casas}, \binits{D.d.l.}},
\oauthor{\bsnm{Bressand}, \binits{F.}},
\oauthor{\bsnm{Lengyel}, \binits{G.}},
\oauthor{\bsnm{Lample}, \binits{G.}},
\oauthor{\bsnm{Saulnier}, \binits{L.}},
\oauthor{\bsnm{Lavaud}, \binits{L.R.}},
\oauthor{\bsnm{Lachaux}, \binits{M.-A.}},
\oauthor{\bsnm{Stock}, \binits{P.}},
\oauthor{\bsnm{Scao}, \binits{T.L.}},
\oauthor{\bsnm{Lavril}, \binits{T.}},
\oauthor{\bsnm{Wang}, \binits{T.}},
\oauthor{\bsnm{Lacroix}, \binits{T.}},
\oauthor{\bsnm{Sayed}, \binits{W.E.}}:
{Mistral 7B}
(2023)
\end{botherref}
\endbibitem

\bibitem[\protect\citeauthoryear{Nguyen et~al.}{2024}]{min_p}
\begin{botherref}
\oauthor{\bsnm{Nguyen}, \binits{M.}},
\oauthor{\bsnm{Baker}, \binits{A.}},
\oauthor{\bsnm{Neo}, \binits{C.}},
\oauthor{\bsnm{Roush}, \binits{A.}},
\oauthor{\bsnm{Kirsch}, \binits{A.}},
\oauthor{\bsnm{Shwartz-Ziv}, \binits{R.}}:
{Turning Up the Heat: Min-p Sampling for Creative and Coherent LLM Outputs}
(2024)
\end{botherref}
\endbibitem

\bibitem[\protect\citeauthoryear{Steinmetz and Sattler}{2021}]{KGQA_benchmarks}
\begin{barticle}
\bauthor{\bsnm{Steinmetz}, \binits{N.}},
\bauthor{\bsnm{Sattler}, \binits{K.U.}}:
\batitle{{What is in the KGQA Benchmark Datasets? Survey on Challenges in Datasets for Question Answering on Knowledge Graphs}}.
\bjtitle{Journal on Data Semantics}
\bvolume{10}(\bissue{3-4}),
\bfpage{241}--\blpage{265}
(\byear{2021})
\doiurl{10.1007/s13740-021-00128-9}
\end{barticle}
\endbibitem

\bibitem[\protect\citeauthoryear{Guo et~al.}{2023}]{Guo2023}
\begin{botherref}
\oauthor{\bsnm{Guo}, \binits{Z.}},
\oauthor{\bsnm{Jin}, \binits{R.}},
\oauthor{\bsnm{Liu}, \binits{C.}},
\oauthor{\bsnm{Huang}, \binits{Y.}},
\oauthor{\bsnm{Shi}, \binits{D.}},
\oauthor{\bsnm{{Supryadi}}},
\oauthor{\bsnm{Yu}, \binits{L.}},
\oauthor{\bsnm{Liu}, \binits{Y.}},
\oauthor{\bsnm{Li}, \binits{J.}},
\oauthor{\bsnm{Xiong}, \binits{B.}},
\oauthor{\bsnm{Xiong}, \binits{D.}}:
{Evaluating Large Language Models: A Comprehensive Survey}
(2023)
\end{botherref}
\endbibitem

\end{thebibliography}

\end{document}